\documentclass[fleqn,10pt]{wlscirep}
\usepackage[utf8]{inputenc}
\usepackage[T1]{fontenc}
\usepackage{comment}
\usepackage{braket}
\usepackage{xcolor}
\usepackage{float}

\title{Modeling Eye Gaze Velocity Trajectories using GANs with Spectral Loss for Enhanced Fidelity}
\author[1,2,3,*]{Shailendra Bhandari} %\orcidID{0000-0002-7860-4854} 
\author[1,2,3]{Pedro Lencastre}
\author[1,2]{Rujeena Mathema}
\author[4]{Alex Szorkovszky} %\orcidID{0000-0003-4696-9872} 
\author[1,2,3]{Anis Yazidi} %\orcidID{0000-0003-4696-9872} 
\author[1,2,3,4]{Pedro G. Lind} %\orcidID{0000-0002-8176-666X} }
\affil[1]{Department of Computer Science, OsloMet -- Oslo Metropolitan University, P.O.~Box 4 St.~Olavs plass, N-0130 Oslo, Norway}
\affil[2]{OsloMet Artificial Intelligence Lab, Pilestredet 52, N-0166 Oslo, Norway}
\affil[3]{{\it NordSTAR} -- Nordic Center for Sustainable and Trustworthy AI Research, Pilestredet 52, N-0166 Oslo, Norway}

\affil[4]{Simula Research Laboratory, Numerical Analysis and Scientific Computing, Oslo, 0164, Norway}
\affil[*]{shailendra.bhandari@oslomet.no}

\begin{abstract}

Accurate modeling of eye gaze dynamics is essential for advancement in human-computer interaction, neurological diagnostics, and cognitive research. Traditional generative models like Markov models often fail to capture the complex temporal dependencies and distributional nuance inherent in eye gaze trajectories data. This study introduces a Generative Adversarial Network (GAN) framework employing Long Short-Term Memory (LSTM) and Convolutional Neural Network (CNN) generators and discriminators to generate high-fidelity synthetic eye gaze velocity trajectories. We conducted a comprehensive evaluation of four GAN architectures: CNN-CNN, LSTM-CNN, CNN-LSTM, and LSTM-LSTM—trained under two conditions: using only adversarial loss (\(L_G\)) and using a weighted combination of adversarial and spectral losses.
  Our findings reveal that the LSTM-CNN architecture trained with this new loss function %(\( L_{\text{final}} \)) 
  exhibits the closest alignment to the real data distribution, effectively capturing both the distribution tails and the intricate temporal dependencies. The inclusion of spectral regularization significantly enhances the GANs' ability to replicate the spectral characteristics of eye gaze movements, leading to a more stable learning process and improved data fidelity. Comparative analysis with an HMM optimized to four hidden states further highlights the advantages of the LSTM-CNN GAN. Statistical metrics show that the HMM-generated data significantly diverges from the real data in terms of mean, standard deviation, skewness, and kurtosis. In contrast, the LSTM-CNN model closely matches the real data across these statistics, affirming its capacity to model the complexity of eye gaze dynamics effectively. These results position the spectrally regularized LSTM-CNN GAN as a robust tool for generating synthetic eye gaze velocity data with high fidelity. Its ability to accurately replicate both the distributional and temporal properties of real data holds significant potential for applications in simulation environments, training systems, and the development of advanced eye-tracking technologies, ultimately contributing to more naturalistic and responsive human-computer interactions.

\end{abstract}

\begin{document}
%%%%%%%%%%%%%%%%%%%%%%%%%%%%%%%%%%%%%%%%%%%%%%%%%%%%%%%%%%%%%%%%%%%%%%%%%%%%%%%%%%%%%%%%%%%%%%%%%%%

\flushbottom
\maketitle
\thispagestyle{empty}

%%%%%%%%%%%%%%%%%%%%%%%%%%%%%%%%%%%%%%%%%%%%%%%%%%%%%%%%%%%%%%%%%%%%%%%%%%%%%%%%%%%%%%%%%%%%%%%%%%%%%%%%%%%%%%%
%%%%%%%%%%%%%%%%%%%%%%%%%%%%%%%%%%%%%%%%%%%%%%%%%%%%%%%%%%%%%%%%%%%%%%%%%%%%%%%%%%%%
\section{Introduction}
Eye gaze trajectories provide critical insights into human visual attention, perception, and cognitive processes \cite{10.1167/11.5.5, 000351929300021}. Modeling the dynamics of stochastic processes, such as eye gaze trajectories, is a significant challenge with practical applications in fields like human-computer interaction, cognitive science, and artificial intelligence (AI). These trajectories are critical for understanding behaviors in tasks like reading, visual search, and interaction with computer systems \cite{shapiro2001computer}. The sequential nature and inherent stochasticity of eye movement patterns pose significant challenges for accurate modeling and prediction. Capturing these dynamics is not only important for advancing theoretical understanding but also has practical applications in areas like user interface design, psychological assessment, and marketing research \cite{Ball2022}. Traditional models of eye movement, such as Markov models, offer approximate estimations by assuming that future states depend solely on the present state \cite{PATTERSON200887}. However, these models often fall short in accounting for the complex interdependencies influenced by perceptual constraints, memory, and cognitive factors. Comparisons have been drawn between visual search patterns and animal foraging behaviors, highlighting the utility of stochastic models in capturing the nuances of eye movement \cite{Rhodes2014}. Visual search tasks involve goal-directed scanning patterns while foraging behaviors resemble exploratory gaze movements, both reflecting distinct cognitive strategies. Studies by Kerster et.al. \cite{KERSTER201685} have compared these behaviors, revealing that human gaze patterns can exhibit characteristics akin to animal foraging in natural environments.

Our objective here is to develop models that accurately capture the sequential nature of eye gaze velocities during a visual search task. By representing these velocities as stochastic processes, we aim to enhance the understanding of eye movement dynamics and improve the generation of synthetic eye-tracking data. For this purpose, we explore the Generative Adversarial Networks (GANs), and Markov models.  The spatial dynamics of eye-gaze trajectories are typically modeled as time-continuous or discrete stochastic processes \( (X_t)_{t \geq 0} \), with \( t \) denoting time. Under the classical Markovian assumption, the future state distribution of eye-gaze trajectories can be completely predicted given its present state, a premise discussed in depth for animal movement by Patterson et al.\ \cite{PATTERSON200887}. HMMs are powerful statistical tools designed to model sequences where the system being modeled is assumed to be a Markov process with unobservable (hidden) states \cite{Awad2015}. They have proven highly effective in fields such as animal movement modeling \cite{langrock2012flexible, WOS:000331402000011, 000351929300021, https://doi.org/10.1111/2041-210X.12657} due to their efficiency in fitting discrete latent state models to time series data. The foundational implementation of HMMs in ecological modeling is comprehensively explored by McClintock and Michelot \cite{https://doi.org/10.1002/ecy.1880}, which also provides the source code for replication and further analysis.  However, Markov models provide only an approximate estimation of eye movement, as they do not account for intricate interdependencies due to perceptual constraints, memory, and cognitive factors influencing human vision.

Recent advancements in artificial intelligence (AI) and deep learning have significantly propelled the field of generative models \cite{Harshavardha2020,Oussidi2018}. These models aim to learn the underlying probability distribution of data to generate new samples with properties similar to the training data. Common approaches include variational autoencoders (VAEs) \cite{Pu2016VariationalAF}, generative adversarial networks (GANs) \cite{goodfellow2014generative,10.1145/3422622}, and deep reinforcement learning (DRL) \cite{Harshavardha2020}. Among these methods, GANs have achieved remarkable success and have outperformed other models in various tasks \cite{Gonog2019,Harshavardha2020,Laptev2021}. The core concept of GANs involves two adversarial networks—the generator and the discriminator—in a supervised learning framework \cite{goodfellow2014generative}. The generator attempts to create fake samples to deceive the discriminator, which learns to distinguish between real and generated samples based on the real data distribution. Research on GANs has expanded rapidly in recent years, with their primary impact observed in the field of computer vision \cite{brock2019largescalegantraining,Pathak2016ContextEF}, particularly in creating realistic images and videos. While these developments have attracted considerable attention, GAN applications have also broadened into other areas, including time series and sequence generation \cite{10.1145/3559540}. As a relatively new area of exploration, ongoing research efforts aim to generate high-quality, diverse, and privacy-preserving time series data using GANs. Foundational methods like VAEs and GANs aim to approximate a latent-space model of the underlying distributions of images or time series based on training data samples. With this latent-space model, it is possible to generate new synthetic samples and alter their semantic properties across different dimensions. Developing a well-performing GAN model involves two phases \cite{Shmelkov2018}: training the generator and discriminator using real training samples, and validating the generated outputs against a reference test set of real samples. The main objective during training is to measure and minimize the divergence between the probability distributions of the generated data and the real data using a quantifiable objective function \cite{Cai2020,NIPS2014_5ca3e9b1}. Despite the remarkable progress and intriguing potential of GANs \cite{PavanKumar2020GenerativeAN}, they encounter significant challenges, including unstable training processes \cite{10386654}, greater computational demands compared to other deep learning methods \cite{orponen1994computational}, and issues such as vanishing gradients and mode collapse \cite{arjovsky2017wasserstein}. Both academic and industrial efforts have been directed toward addressing these challenges \cite{garg2020advances,wiebe2015quantum}%,PhysRevResearch.2.033212,doi:10.1126/sciadv.aav2761,stein2022quclassi}.

%We evaluated four different GAN architectures corresponding to every generator and discriminator pair for the considered 1D-CNN and LSTM architectures as shown in Fig.~\ref{fig:GAN_architecture}. 
This study focuses on evaluating the capability of GANs, and the HMM to replicate stochastic trajectories and compare their performance against established stochastic modeling benchmarks. Previous studies, such as Lencastre et al. \cite{LENCASTRE2023133831}, have contrasted various GAN architectures- including Recurrent Conditional GAN (RCGAN), Time-series GAN (TimeGAN), Signature Conditional Wasserstein GAN (SigCWGAN), and Recurrent Conditional Wasserstein GAN (RCWGAN)—with Markov chain models. These studies observed that GANs often struggle to capture rare events, and cross-feature relationships, and produce accurate synthetic data due to their inability to effectively model the complex temporal dependencies and distributional nuances inherent in eye-gaze trajectories. Notably, simpler models like Markov models have often outperformed these GANs in such tasks. In this analysis, we address these limitations by incorporating spectral loss into GAN training and employing lightweight LSTM and CNN architectures.  By adding spectral loss, we enhance the GAN's ability to capture both the global and local frequency characteristics of eye-gaze trajectories, leading to more accurate replication of complex temporal dependencies and distributional nuances. This addition stabilizes the training process and significantly improves modeling fidelity, effectively bridging the performance gap identified in prior studies.  We investigate the performance of four classical GAN configurations—CNN-CNN, CNN-LSTM, LSTM-CNN, and LSTM-LSTM. Our spectrally regularized GAN significantly improves the modeling fidelity, effectively capturing both the distribution tails and intricate temporal dependencies of eye-gaze data. By focusing on balancing accuracy with computational efficiency, our approach provides a practical solution without the need for more complex architectures. This approach allowed us to evaluate the effectiveness of different network architectures in capturing the temporal dynamics of eye-gaze trajectories and generating high-quality synthetic data. While we acknowledge the existence of various advanced GAN methods for time-series data, comprehensive comparisons with all existing GAN models are beyond the scope of this paper but represent important future work.

%%%%%%%%%%%%%%%%%%%%%%%%%%%%%%%%%%%%%%%%%%%%%%%%%%%%%%%%%%%%%%%%%%%%%%%%%%%%%%%%%%%%%%%%%%%%%%%%%%%%%%%%%%%%%%%%%%%%%%%%%%%%%%%%%%%%%%%%%%%%%%%%%%%%%%%%%%%%%%%%%%%%%%%%%%%%%%%%%%%%%%%%%%%%%%%%%%%%%%%%%%%%%%%%%%METHODS%%%%%%%%%%%%%%%%%%%%%%%%%%%%%%%%%%%%%%%%%%%%%%%%%%%%%%%%%%%%%%%%%%%%%%%%%%%%%%%%%%%%%%%%%%%%%%%%%%%%%%%%%%%%%
\section{Methods}
\subsection{Generative Adversarial Networks}
A variety of probabilistic models describe animal movement by considering the position as an output of a deterministic function \( G \) applied to a sampler of random latent variables. For example, in correlated random walks (CRWs), let \( s_t \) and \( \phi_t \) represent the step length and turning angle at time \( t \), then the CRW model can be formulated as \cite{PATTERSON200887}:

\begin{equation}
\begin{pmatrix}
(s_1, \phi_1) \\
\vdots \\
(s_n, \phi_n)
\end{pmatrix}
=
\begin{pmatrix}
\left(F^{-1}(Z_G^1 \mid \theta_F),\; H^{-1}(Z_H^1 \mid \phi_0, \theta_H) \right) \\
\vdots \\
\left(F^{-1}(Z_G^n \mid \theta_F),\; H^{-1}(Z_H^n \mid \phi_{n-1}, \theta_H) \right)
\end{pmatrix}
= G(z),
\end{equation}
where \( F \) and \( H \) are cumulative distribution functions (CDFs) with parameters \( \theta_F \) and \( \theta_H \), typically derived from log-normal and von Mises probability density functions, respectively. The variables \( Z_G^n \) and \( Z_H^n \) are random variables representing the latent space, and \( z \) denotes the collection of these latent variables. The generative aspect of GANs applies the deterministic function \( G \) to latent variables sampled according to a specified distribution. GANs comprise two neural networks—the generator \( G \) and the discriminator \( D \)—trained concurrently. The discriminator learns to differentiate between data generated by \( G(z) \) and real data, enhancing the generator's ability to effectively replicate the empirical data distribution.

The GAN framework we are using in this work consists of these two neural networks: the generator and the discriminator, as shown in Fig.~\ref{fig:GAN_architecture}. They are trained alternately. Consider a classical training dataset \( X = \{x^0, x^1, \dots, x^{s-1}\} \) drawn from an unknown time series distribution. The generator \( G(z) \) receives a random noise vector \( z \) sampled from a prior distribution \( P_z(z) \) and produces the generated sample \( G(z) \). The discriminator \( D \) is trained to distinguish between the training data \( x \) and the generated data \( G(z) \). The parameters of \( D \) are updated in order to maximize \cite{SITU2020193}:
\begin{equation}
\label{eq:discriminator_loss}
\mathbb{E}_{x \sim P_{d}(x)}\left[\log D(x)\right] + \mathbb{E}_{z \sim P_{z}(z)}\left[\log\left(1 - D\left(G(z)\right)\right)\right],
\end{equation}
where \( P_d(x) \) is the real time series distribution and \( P_z(z) \) is the distribution of the input noise. The output \( D(x) \) represents the probability that \( D \) classifies the sample \( x \) as real. The goal of the discriminator is to maximize the probability of correctly classifying real and generated samples, making \( D \) a better adversary so that \( G \) must improve to fool \( D \).
Similarly, the parameters of the generator \( G \) are updated to maximize:
\begin{equation}
\label{eq:generator_loss}
\mathbb{E}_{z \sim P_{z}(z)}\left[\log D\left(G(z)\right)\right],
\end{equation}
to convince \( D \) that the generated samples \( G(z) \) are real.
The goal of optimizing classical GANs can be approached from several perspectives. In this study, we adopt the non-saturating loss function \cite{fedus2018paths}, which is also implemented in the original GAN publication's code \cite{NIPS2014_5ca3e9b1}. The generator loss function is given by:
\begin{equation}
\label{eq:generator_loss_ns}
L_G = -\mathbb{E}_{z \sim P_{z}(z)}\left[\log D\left(G(z)\right)\right],
\end{equation}
which aims to maximize the likelihood that the generator creates samples labeled as real data samples. In addition, the discriminator's loss function is given by:
\begin{equation}
\label{eq:discriminator_loss_ns}
L_D = \mathbb{E}_{x \sim P_{d}(x)}\left[\log D(x)\right] + \mathbb{E}_{z \sim P_{z}(z)}\left[\log\left(1 - D\left(G(z)\right)\right)\right],
\end{equation}
which aims at maximizing the likelihood that the discriminator labels real data samples as real and generated data samples as fake. The expected values are approximated in practice using mini-batches of size \( m \). The generator's loss becomes:
\begin{equation}\label{eq:lossG}
L_G = -\frac{1}{m} \sum_{i=1}^{m} \log D\left(G\left(z^{(i)}\right)\right),
\end{equation}
and the discriminator's loss is:
\begin{equation}\label{eq:lossD}
L_D = \frac{1}{m} \sum_{i=1}^{m} \left[ \log D\left(x^{(i)}\right) + \log \left(1 - D\left(G\left(z^{(i)}\right)\right)\right) \right],
\end{equation}
where \( x^{(i)} \) are samples from the real dataset \( X \) and \( z^{(i)} \) are noise samples from \( P_z(z) \).

\subsection{Spectral Loss}
To enhance the training of the generator, we complement its loss function with additional terms, including an application-specific term and a regularization term known as the spectral loss \( L_{\text{spectral}} \) \cite{8099502,Durall_2020_CVPR}. Spectral loss is a novel loss function designed to enhance generative models by embedding frequency domain insights into the training framework. It leverages the Fourier transform \( F \), which decomposes a signal into its constituent frequencies, to compare the real data sequence \( (x_0, x_1, \ldots, x_n) \) with the generated sequence \( (\hat{x}_0, \hat{x}_1, \ldots, \hat{x}_n) \) \cite{https://doi.org/10.1111/2041-210X.13853}.
We define the spectral loss incorporated into the generator's gradient descent as:
\begin{equation}\label{eq:finalloss}
    L_{\text{spectral}} = \sum_{k=0}^{N-1} \left[ \log\left( \left| F(x)_k \right| \right) - \log\left( \left| F(\hat{x})_k \right| \right) \right]^2,
\end{equation}
where \( |F(x)_k| \) is the magnitude of the Fourier transform of the real data sequence at frequency \( k \), and similarly for \( F(\hat{x})_k \). Here, \( x \) and \( \hat{x} \) are the real and generated velocity trajectories, respectively.
The discrete Fourier transform \( F \) of a one-dimensional time series \( x \) of length \( N \) is defined as:
\begin{equation}
F(x)_k = \sum_{n=0}^{N-1} x_n \cdot e^{-2\pi i \frac{kn}{N}},
\end{equation}
for \( k = 0, 1, \ldots, N-1 \).
By applying the logarithm to the magnitude of the Fourier transforms, the loss function normalizes amplitude disparities, focusing on the relative spectral energy distribution. The squared term captures the spectral discrepancy between real and synthetic data, assigning higher penalties to more significant differences. The summation over all frequencies consolidates the error across the frequency spectrum into a single scalar value, quantifying the generated data's spectral fidelity.
%%%%%%%%%%%%%%%%%%%%%%%%%%%%%%%%%%%%%%%%%%%%%%%%%%%%%%%%%%%%%%%%%%%%%%%%%%%%%%%%%%%%%%%%%%%%%%%%%%%%%%%%%%%%%%%%%%%%%%%%%%%%%%%%%%%%%%%%%%%%%%%%%%%%%%%%%%%%%%%%%%%%%%%%%%%%%%%%%%%%%%%%%%%%%%%%%%%%%%%%%%%%%%%%%%%%%%%%%%%%%%%%%%%%%%%%%%%%%%%%%%%%%%%%%%%%%%%%%%%%%%%%%%%%%%%%%%%%%%%%%%%%%%%%%%%%%%%%%%%%%%%%%%%%%%%%%%%%%%%%%%%%%%%%%%%%%%%%%%%
\begin{figure}[t]
    \centering
    \includegraphics[width=\linewidth]{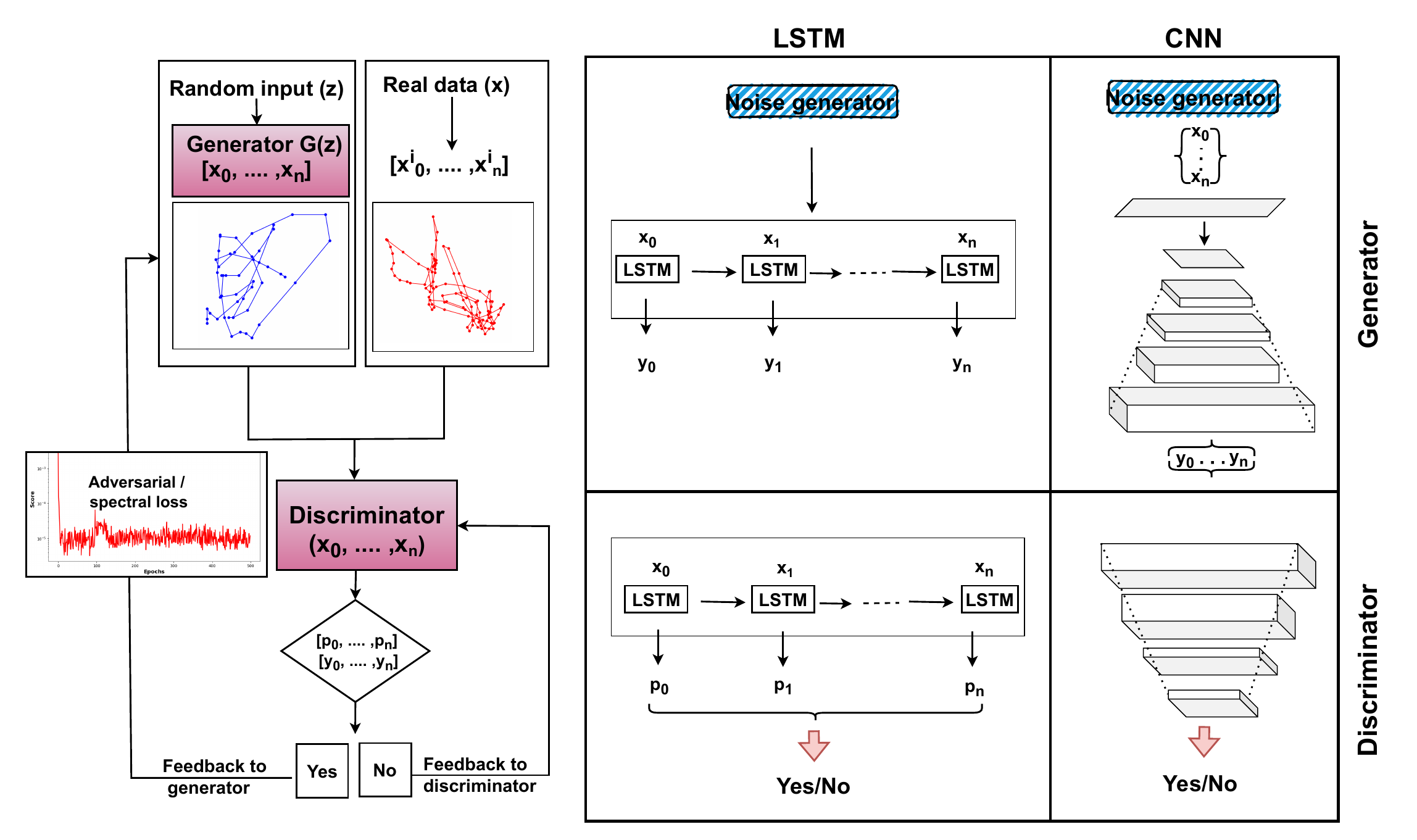}
    \caption{Overview of the GAN process for generating eye-tracking data, showing the interaction between the generator, discriminator, and adversarial/spectral loss feedback. Detailed architectures of LSTM and CNN-based generators and discriminators, illustrating possible combinations for generating and discriminating gaze trajectory data from noise input.}
    \label{fig:GAN_architecture}
    %\Description{GAN Architecture.}

\end{figure}
%%%%%%%%%%%%%%%%%%%%%%%%%%%%%%%%%%%%%%%%%%%%%%%%%%%%%%%%%%%%%%%%%%%%%%%%%%%%%%%%%%%%%%%%%%%%%%%%%%%%%%%%%%%%%%%%%%%%%%%%%%%%%%%%%%%%%%%%%%%%%%%%%%%%%%%%%%%%%%%%%%%%%%%%%%%%%%%%%%%%%%%%%%%%%%%%%%%%%%%%%%%%%%%%%%%%%%%%%%%%%%%%%%%%%%%%%%%%%%%%%%%%%%%%%%%%%%%%%%%%%%%%%%%%%%%%%%%%%%%%%%%%%%%%%%%%%%%%%%%%%%%%%%%%%%%%%%%%%%%%%%%%%%%%%%%%%%%%%%%
Spectral regularization is then defined by combining the generator loss \( L_G \) with the spectral loss \( L_{\text{spectral}} \):
\begin{equation}
    L_{\text{final}} = L_G + \lambda L_{\text{spectral}},
\end{equation}
where \( \lambda \) is the hyperparameter that weights the influence of the spectral loss.
Various deep-learning architectures can be employed for the generator and discriminator networks in GANs. Among the most popular, efficient, and widely used techniques are Long Short-Term Memory (LSTM) networks and Convolutional Neural Networks (CNNs) \cite{alom2019}. LSTMs are designed with memory cells that capture temporal dependencies in sequential data, making them highly suitable for modeling the temporal patterns inherent in eye-tracking datasets. CNN, on the other hand, excels at capturing spatial and spatiotemporal features through convolutional layers, which is advantageous for analyzing the spatial characteristics of eye-tracking data \cite{https://doi.org/10.1155/2019/8641074}. In all models, training utilized vectors of length 256 with random noise uniformly distributed between 0 and 1 as input. The models underwent 500 epochs of training with a learning rate of 0.0002, optimizing the combined loss function described above. To assess performance, we calculated the mean squared error between the predicted and real logarithmic Fourier decomposition spectra, as defined by Equations~\eqref{eq:lossG}, \eqref{eq:lossD}, and \eqref{eq:finalloss}. In this study, we utilize two architectures for both the generator and discriminator: CNN-based and LSTM-based architectures (see Fig.~\ref{fig:GAN_architecture}). Below, we provide a concise overview of the purpose and functionality of these networks. For a more comprehensive introduction to deep networks, we direct the reader to Christin et al.\cite{https://doi.org/10.1111/2041-210X.13256}.

%%%%%%%%%%%%%%%%%%%%%%%%%%%%%%%%%%%%%%%%%%%%%%%%%%%%%%%%%%%%%%%%%%%%%%%%%%%%%%%%%%%%%%%%
\subsection{Long Short-Term Memory (LSTM) Networks and Convolutional Neural Networks (CNNs)}

LSTM networks are sophisticated architectures within recurrent neural networks (RNNs), particularly adept at modeling time series data, including trajectory analyses. A distinctive feature of LSTMs is their ability to capture and leverage long-term dependencies using gating mechanisms \cite{alom2019}. These architectures are widely employed in GANs for tasks such as predicting pedestrian movements and generating medical time-series data \cite{esteban2017realvalued,10.1145/3474838}. In our study, the generator network incorporates an LSTM layer specifically configured to accept a unique random seed at each time step. This LSTM layer generates a sequence of hidden vectors, each consisting of 16 attributes that capture the state of the eye-gaze trajectory. Subsequently, a dense layer processes each 16-dimensional hidden vector at predetermined time intervals, converting them into corresponding horizontal and vertical displacements. These displacements, when divided by the time interval, represent the velocity components of the eye-gaze trajectory. By aggregating these incremental displacements, we compile a detailed time series of the eye-gaze velocity, as illustrated in Fig.~\ref{fig:GAN_architecture}. The LSTM architecture is also employed in our discriminator. This component processes a sequence of positions, symbolically representing points in the eye-gaze path rather than actual gaze coordinates. The LSTM transforms these positions into a higher-dimensional latent space. A subsequent dense layer assesses each point in the sequence for its probability of being a plausible part of a trajectory. The discriminator's output is a mean probability value, which is used to evaluate the accuracy and realism of the predicted eye-gaze trajectory.

CNN architectures utilize convolutional layers and are at the forefront of technologies for a multitude of applications, notably in signal and image processing. They excel at extracting both low-level and high-level features from multidimensional tensors \cite{alom2019}. In GANs, CNNs are extensively utilized \cite{Radford2015UnsupervisedRL}. Our approach adopts the architecture proposed by Radford et al. \cite{Radford2015UnsupervisedRL} for image generation tasks. The generator begins with a random noise vector, which serves as a latent representation of a comprehensive time series. This is followed by a sequence of fractional-strided convolutions that progressively transform the latent representation into a time series with an increasing number of points and a decreasing number of features, culminating in a two-dimensional vector of the specified length (refer to Fig.~\ref{fig:GAN_architecture}). In our implementation, batch normalization and ReLU activations are applied after each fractional-strided convolution, except for the final output layer, which employs a hyperbolic tangent activation as recommended by Radford et al. \cite{Radford2015UnsupervisedRL}. It is important to note that this CNN framework does not perform explicit sequential modeling of trajectories, and the latent representations are not necessarily time-correlated. For the CNN-based discriminator, we employ a series of strided convolutions to progressively convert the initial trajectory into a time series characterized by shorter lengths and more features, eventually producing a latent vector that encapsulates the entire trajectory. Batch normalization and LeakyReLU activations enhance the functionality of each convolution. The culmination of this process is a dense layer equipped with a sigmoid activation, which translates the latent vector into a probability assessing the realism of the trajectory (see Fig.~\ref{fig:GAN_architecture}). The hyperparameters of the model architecture are listed in Table \ref{tab:hyperparameters}. 
%%%%%%%%%%%%%%%%%%%%%%%%%%%%%%%%%%%%%%%%%%%%%%%%%%%%%%%%%%%%%%%%%%%%%%%%%%%%%%%%%%%%%%%%%%%%%%%%%%%%%%%%%%%%%%%%%%%%%%%%%%%%%%%%%%%%%%%%%%%%%%

\begin{table}[t]
\caption{Summary of hyperparameters for the GAN model.}
\centering
\begin{tabular}{lcc}
\toprule
\textbf{Hyperparameter} & \textbf{Value} \\
\midrule
Sequence length & 200  \\
Batch size & 128  \\
Learning rate & 0.0002  \\
Optimizer & Adam  \\
$\beta_1$, $\beta_2$ & 0.5, 0.999  \\
Input channels & 256  \\
Epochs & 500  \\
Spectral loss ($L_{spectral}$) & Yes  \\
$\lambda$ & 0.1\\
\bottomrule
\end{tabular}
\label{tab:hyperparameters}
\end{table}

%%%%%%%%%%%%%%%%%%%%%%%%%%%%%%%%%%%%%%%%%%%%%%%%%%%%%%%%%%%%%%%%%%%%%%%%%%%%%%%%%%%%%%%%
\subsection{Markov Models}
This study employs Hidden Markov Models (HMMs) to model and analyze eye-gaze velocity time series data. Markov models are statistical frameworks capturing dependencies between current and recent states (typically, the most recent) in a time series, enabling the generation of synthetic data that preserves the temporal patterns observed in the original dataset \cite{Bremaud2020}. 
%A high-order Markov model is used to estimate transition probabilities with kernel density estimation (KDE). This method effectively captures temporal dependencies up to a specified lag \( k \), generating data that maintains the statistical properties of the original time series. 
Hidden Markov Models extend this concept by incorporating unobserved (hidden) states, making them powerful for modeling systems where the underlying process is not directly observable \cite{rabiner1989tutorial}. In the context of eye-gaze data, hidden states may represent different types of eye movements, such as fixations and saccades. At the same time, the observations (known as emissions) correspond to measured eye-gaze velocities. We used the \emph{Baum-Welch algorithm} \cite{baum1970maximization}, a specialized Expectation-Maximization (EM) algorithm, for parameter estimation. This iterative algorithm refines estimates of the initial state probabilities, state transition probabilities, and emission probabilities to maximize the likelihood of the observed data. Utilizing HMMs facilitates the inference of hidden states and the optimization of model parameters based on observed data, supporting a detailed analysis of eye movement behaviors. For detailed mathematical formulations of the Markov and Hidden Markov models employed in this study, refer to Appendix \ref{appexA}.

%%%%%%%%%%%%%%%%%%%%%%%%%%%%%%%%%%%%%%%%%%%%%%
\begin{figure}[t]
    \centering
    \includegraphics[width=1\linewidth]{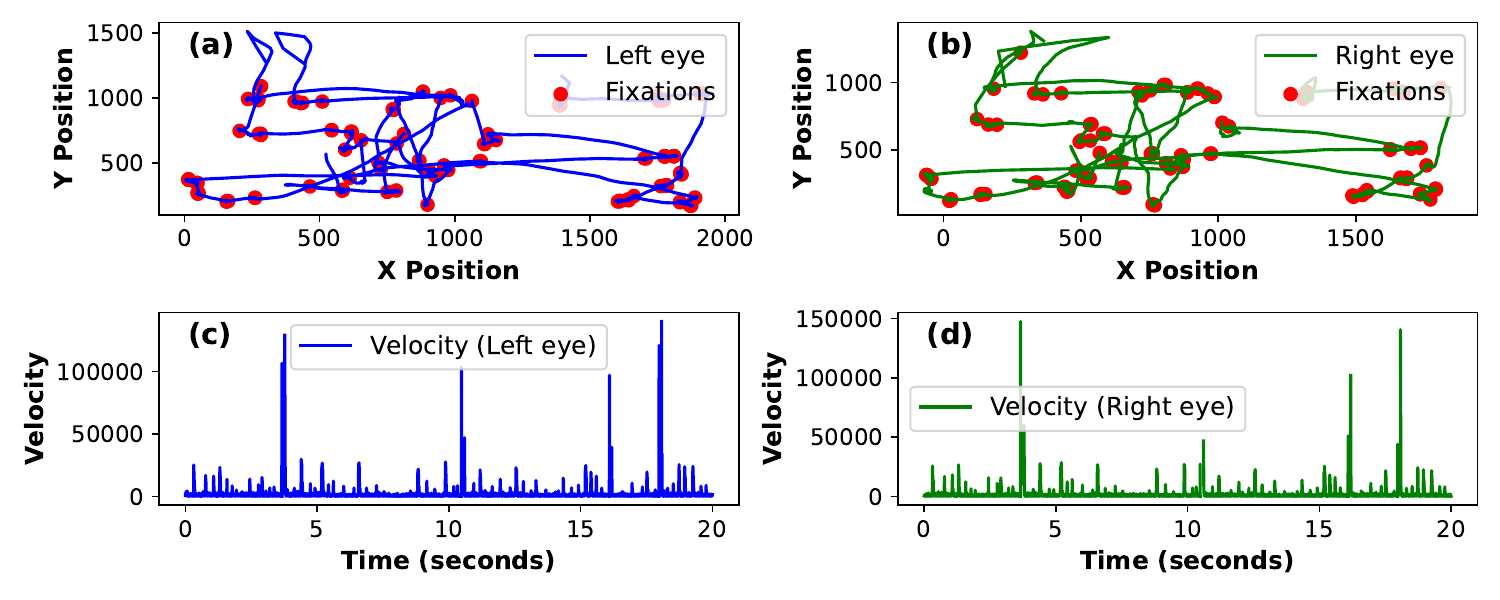}
    \includegraphics[width=1\linewidth]{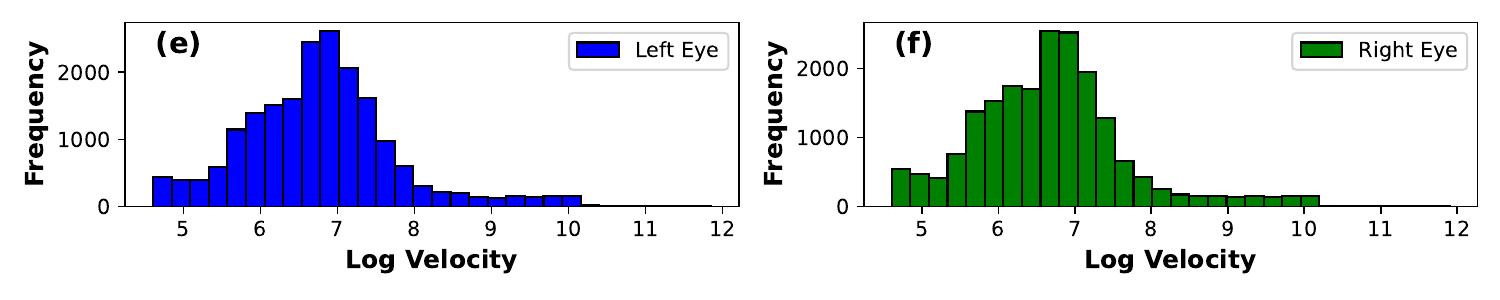}
    \caption{(a, b, c, d) X-Y positions and velocity plots for left and right eye movements. The top two plots show the eye movement trajectories (left and right eyes) in terms of X and Y positions, while the bottom two plots show the respective velocity profiles over time. (e, f) Log-scaled velocity distributions for left and right eye movements for one participant. The histograms represent the distribution of logarithmic velocity values for both eyes.}
    \label{fig:data}
    %\Description{Data.}
\end{figure}
%%%%%%%%%%%%%%%%%%%%%%%%%%%%%%%%%%%%%%%%%%%%%%

%%%%%%%%%%%%%%%%%%%%%%%%%%%%%%%%%%%%%%%%%%%%%%%%%%%%%%%%%%%%%%%%%%%%%%%%%%%%%%%%%%%%%%%%%%%%%%%%%%%%%%%%%%%%%%%%%%%%%%%%%%%%%%%%%%%%%%%%%%%%%%%%%%%%%%%%%%%%%%%%%%%%%%%%%%%%%%%%%%%%%%%%%%%%%%%%%%%%%%%%%%%%%%%%%%%%%%%%%%%%%%%%%%%%%%%%%%%%%%%%%%%%%%%%%%%%%%%%%%%%%%%%
\section{Data and the Statistical Measure}

We used eye-tracking data\footnote{All data collected was anonymized and follows the ethical requirements from the Norwegian Agency for Shared Services in Education and Research (SIKT), under the application with Ref. 129768.} that was gathered at Oslo Metropolitan University utilizing the advanced Eye-Link Duo device \cite{srresearch2024eyelink}, capable of reaching up to 2000 Hz but was adjusted to 1000 Hz for this study. Participants were tasked with searching for specific targets within images from the book Where's Waldo? \cite{bookwaldo}. The measurements were recorded in screen pixels, capturing the nuanced movements of both left and right eyes. Fig.~\ref{fig:data} provides a comprehensive visualization of the collected eye-tracking data. Fig.~\ref{fig:data} (a,b) display the X-Y position trajectories for the left and right eyes, respectively, illustrating the spatial patterns during the search task. The subsequent panels (Fig.~\ref{fig:data} (c,d)) depict the corresponding velocity profiles over time, highlighting the dynamics of eye movements.

The data from eye-tracking measurements were preprocessed and utilized to train a GAN. Initially, the velocity data for both left and right eyes are calculated by finding the Euclidean distance between consecutive position points and then dividing by the time interval, set at one millisecond, to convert this distance into velocity. Fig.~\ref{fig:data} (e,f) presents the log-scaled velocity distributions for one participant's left and right eyes, respectively. These histograms emphasize the range and frequency of velocities encountered during the task. %The velocity data were then structured appropriately for input into the discriminator component of the GAN. 
Subsequently, the dataset was normalized using a MinMaxScaler, scaling the values to fit within the operational range of 0 and 1, as recommended in \cite{DEAMORIM2023109924}. This normalization is crucial for the GAN's training stability and convergence. Finally, the normalized data were segmented into sequences of 200 data points, forming the training batches for the GAN. These sequences were supplied to the discriminator, which learned to differentiate between real eye-tracking data and the synthetic data generated by the generator. This adversarial training process enhanced the discriminator's ability to accurately identify authentic samples, thereby improving the overall performance of the GAN.

The performance of the GAN and HMM model is evaluated using the metric Jensen-Shannon ($D_{JS}$). The $D_{JS}$, a symmetrized version of the Kullback-Leibler (KL) divergence, provides a symmetric measure of distance between probability distributions \cite{e21050485, weng2019gan}.  The JS divergence \cite{weng2019gan} is defined by 
\begin{equation}
D_{JS}(P | Q) = \frac{1}{2} D_{KL}(P | M) + \frac{1}{2} D_{KL}(Q | M),
\end{equation}
where $P$ and $Q$ are distributions, and \( M = \frac{1}{2}(P + Q) \). The KL divergence is standard for assessing distributional similarity, enhancing maximum likelihood estimates. While preserving these properties, the $D_{JS}$ is more intuitive as it assesses the approximation of synthetic distributions to empirical ones. Therefore, $D_{JS}$ can be relevant for discriminators within GANs to distinguish synthetic data from the generator. This measurement is always non-negative and its value is bounded by [0,1] $(0\leq D_{JS}\leq1)$, a lower value indicating higher similarity between the distribution of real and the generated data.

Complementing the $D_{JS}$, the spectral loss \({L}_{\text{spectral}}\) introduces a frequency domain perspective to the evaluation of generative models \cite{8099502, Durall_2020_CVPR}. By utilizing the Fourier transform, the spectral loss compares the frequency content of real and synthetic sequences, focusing on capturing the spectral fidelity of the generated data. The loss function emphasizes the alignment of frequency characteristics between datasets by normalizing amplitude disparities and penalizing significant differences across the frequency spectrum. This approach is particularly advantageous for data exhibiting spatial or temporal patterns, ensuring that the model's output maintains the structural coherence essential for high-quality generation. By integrating the Spectral loss into the generator's optimization process, we not only refine the generator's ability to produce realistic samples but also improve the overall stability and quality of the GAN model's outputs. Together, these metrics provide a comprehensive framework for assessing and enhancing the performance of GANs. Fig.~\ref{fig:modeltradeoff} shows the performance comparison of different GAN models over 500 epochs. The top left panel illustrates the Spectral Score (in log scale), showing how CNN-based models achieve more stable and lower scores compared to LSTM-based models. The top right panel depicts $D_{JS}$ (in log scale), indicating that CNN-based models maintain a closer similarity to the empirical distribution. The bottom panels compare the integral scores (Spectral loss ($L_{final}$) and $D_{JS}$) against the average computation time per epoch, with the LSTM-CNN model achieving the best balance between accuracy and computational efficiency.

When quantifying how well the time dependencies are replicated, we evaluate the autocorrelation of the velocity magnitude \( \|\mathbf{v}\| \). The autocorrelation function (ACF) is defined as:
\begin{equation}
ACF(h) = \frac{\sum_{t=1}^{T-h} (X_t - \bar{X})(X_{t+h} - \bar{X})}{\sum_{t=1}^{T} (X_t - \bar{X})^2},
\end{equation}
where \( X_t \) represents the time series at time \( t \), \( \bar{X} \) is the mean of the series, \( T \) is the total number of observations, and \( h \) is the time lag. The \( ACF(h) \) quantifies the similarity between the data and a shifted version of itself by \( h \) time steps. The values of \( ACF(h) \) range from \(-1\) to \( 1 \), where values closer to \( 1 \) indicate stronger positive correlation at the corresponding lag.

We computed and plotted the autocorrelation coefficient \( ACF(h)_X \) and \( ACF(h)_Y \) between the real and generated data across multiple time lags to evaluate how well the temporal dependencies and patterns of the real data were captured by the GAN-generated data. Although gaze trajectories are not stationary stochastic processes, both the GANs and the HMMs create time-homogeneous trajectories. Therefore, \( ACF(h)_X \) (and \( ACF(h)_Y \), computed similarly) assess the model's ability to replicate the "average" dynamics of the given time series.

%%%%%%%%%%%%%%%%%%%%%%%%%%%%%%%%%%%%%%%%%%%%%%%%%%%%%%%%%%%%%%%%%%%%%%%%%%%%%%%%%%%%%%%%%%%%%%%%%%%%%%%%%%%%%%%%%%%%%%%%%%%%%%%%%%%%%%%%%%%%%%%%%%%%%%%%%%%%%%%%%%%%%%%%%%%%%%%%%%%%%%%%%%%%%%%%%%%%%%%%%%%%%%%%%%%%%%%%%%%%%%%%%%%%%%%%%%%%%%%%%%%%%%%%%%%%%%%%%%%%%%%%%%%%%%%%%%%%%%%%%%%%%%%%%%%%%%%%%%%%%RESULTS%%%%%%%%%%%%%%%%%%%%%%%%%%%%%%%%%%%%%%
%%%%%%%%%%%%%%%%%%%%%%%%%%%%%%%%%%%%%%%%%%%%%%%%%%%%%%%%%%%%%%%%%%%%%%%%%%%%%%%%%%%%%%%%%%%%%%%%%%%%%%%%%%%%%%%%%%%%%%%%%%%%%%%%%%%
\section{Results}
\subsection{Architecture Selection Experiment}
\begin{figure}[t]
    \centering
    \includegraphics[width=0.495\linewidth]{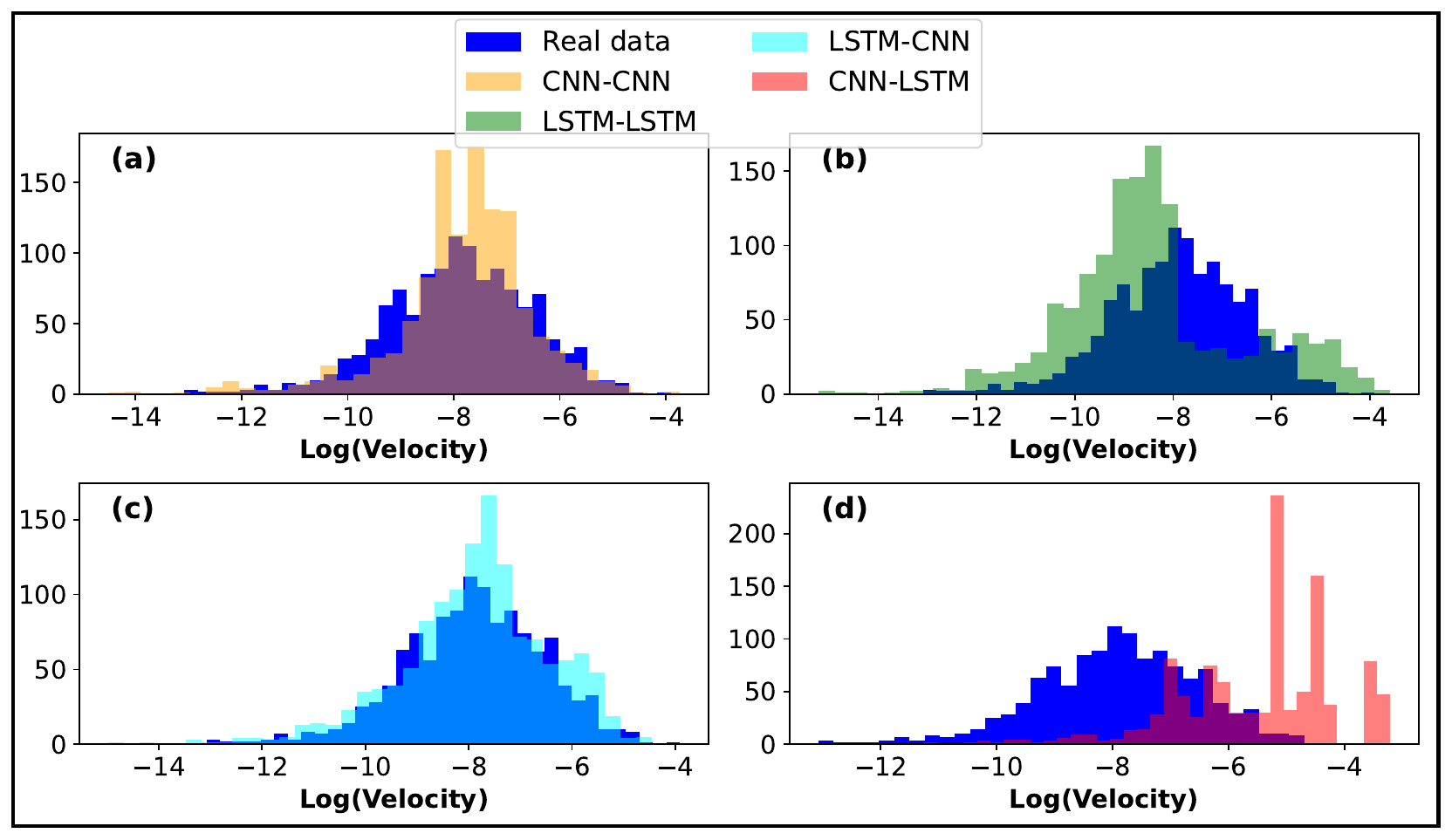}
    \includegraphics[width=0.495\linewidth]{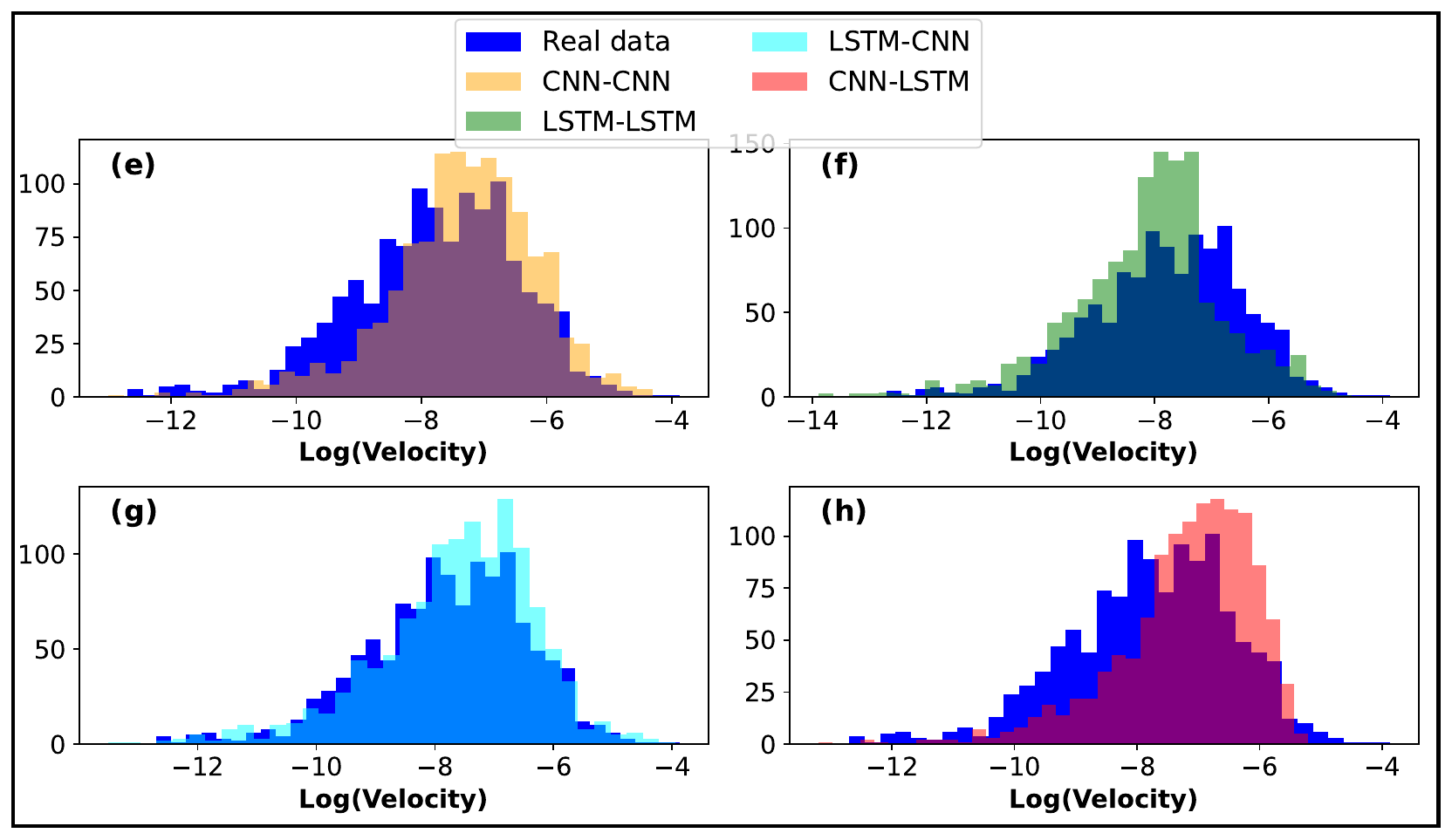}
    \caption{Comparison of velocity distributions for real data (dark blue) and synthetic data generated by GAN models (a-d) with adversarial loss as in Equation \eqref{eq:lossG} and (e-h) with ($L_{\rm final}$) as in Equation \eqref{eq:finalloss}.} 
    \label{fig:PRE_analysis}
    %\Description{Pre Analysis.}
\end{figure}

Fig.~\ref{fig:PRE_analysis} presents log-transformed velocity distributions for real data alongside synthetic data generated by four GAN models: CNN-CNN, LSTM-CNN, CNN-LSTM, and LSTM-LSTM. Each model was trained under two conditions: using only adversarial loss ($L_G$, Fig.~\ref{fig:PRE_analysis} (a-d)) and using adversarial loss combined with spectral loss ($L_{final}$, Fig.~\ref{fig:PRE_analysis} (e-h)). In both conditions, the distributions for real data and generated data are shown, illustrating the alignment of each model's generated data with the real data distribution. Among the models, the LSTM-CNN architectures exhibit the closest alignment to the real data distribution. Furthermore, models trained with $L_{final}$ (right column) show enhanced alignment, particularly in the distribution tails, indicating that spectral regularization ($L_{final}$) helps GAN models capture the true characteristics of the velocity distribution more effectively.

%%%%%%%%%%%%%%%%%%%%%%%%%%%%%%%%%%%%%%%%%%%%%%
%%%%%%%%%%%%%%%%%%%%%%%%%%%%%%%%%%%%%%%%%%%%%%

\begin{figure}[t]
    \centering
    \includegraphics[width=0.53\linewidth]{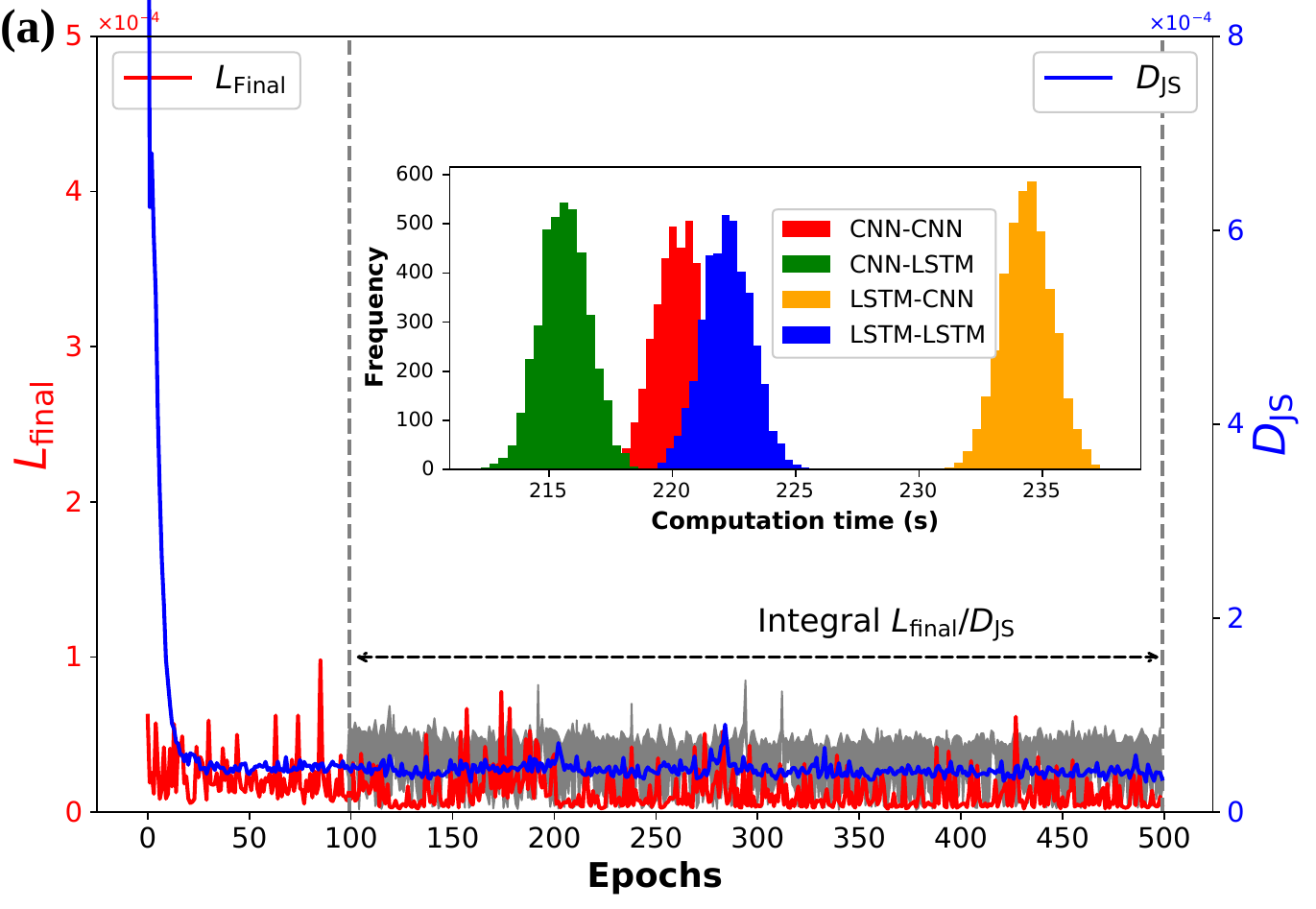}
    \includegraphics[width=0.45\linewidth]{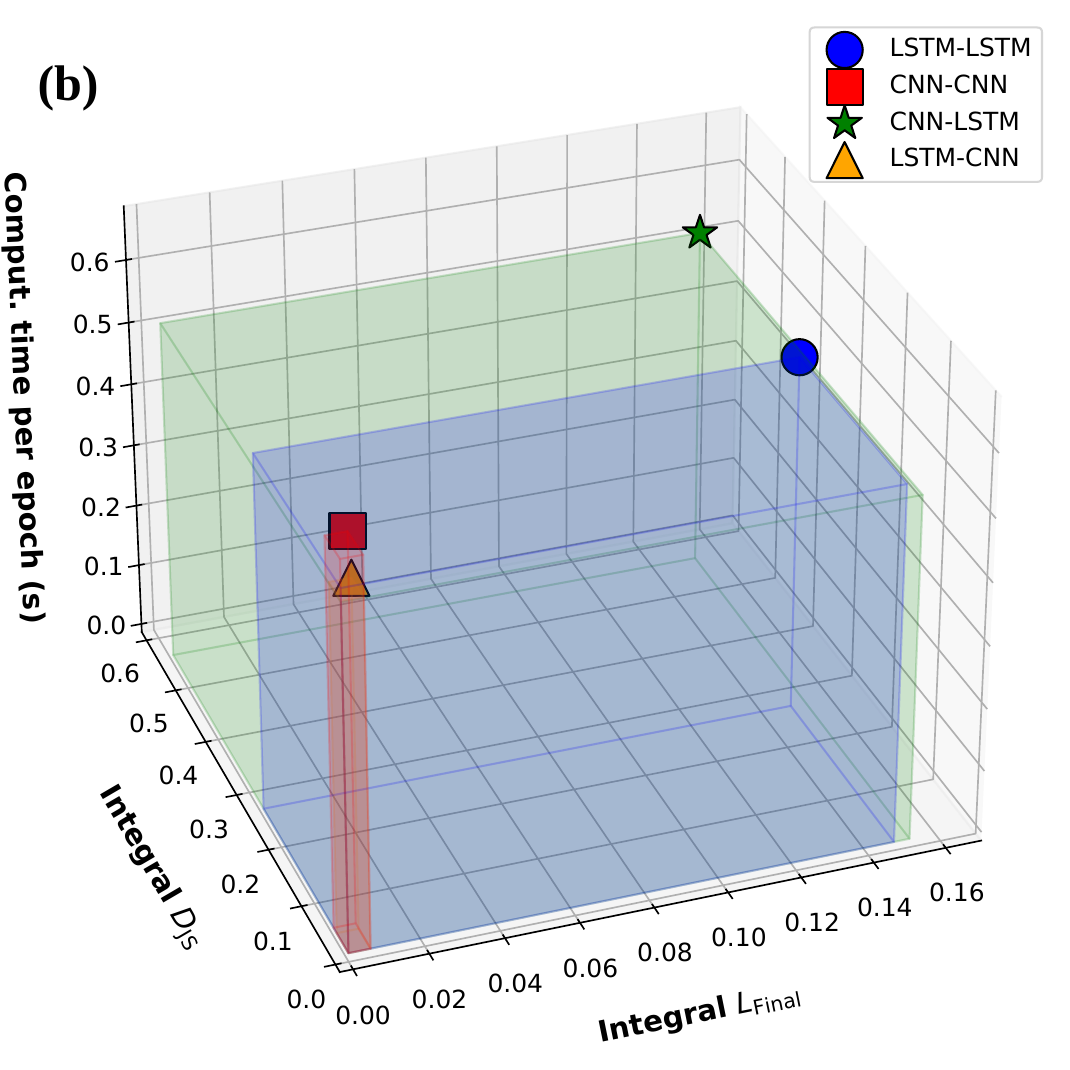}
    \caption{\protect
        %\textcolor{red}{[@PedroLind: revise abstract carefully]} 
        Comparison of neural network architectures (CNN-CNN, LSTM-CNN, CNN-LSTM, LSTM-LSTM) across performance and computational metrics. (a) Total loss (\(L_{\text{final}}\)) over epochs for LSTM-CNN, where \(L_{\text{final}}\) combine GAN loss and spectral loss. The red line represents the spectral loss component (\(L_{\text{spectral}}\)), and the blue line indicates the performance comparison metric \(D_{JS}\), quantifying the similarity between real and generated data. The integral calculation begins at epoch 100 (dashed line). The inset in the same figure is the frequency distribution of computation times across architectures. (b) A 3D representation illustrating the integral values of \(L_{\text{spectral}}\), \(D_{JS}\), and the average computation time per epoch for each model. The LSTM-CNN model demonstrates the most favorable efficiency, achieving the lowest integral volume across loss, performance, and computation time metrics.}
    \label{fig:modeltradeoff}
    %\Description{Results}
\end{figure}

Among the four GAN architectures, marked variations in performance and computational efficiency were observed, as illustrated in Fig.~\ref{fig:modeltradeoff}. Fig.~\ref{fig:modeltradeoff} (a) highlights the progression of the total loss (\(L_{\text{final}}\)) over epochs for the LSTM-CNN architecture. Specifically, the red trajectory corresponds to the spectral loss term (\(L_{\text{spectral}}\)), while the blue trajectory represents the performance metric \(D_{JS}\), which quantifies the alignment between real and generated data distributions. The integral calculation for \(L_{\text{final}}\) and \(D_{JS}\) is applied from epoch 100 to 500 (indicated by the dashed vertical lines). An inset within the same figure further illustrates the frequency distribution of computation times across the architectures, underscoring the computational efficiency of CNN-LSTM and LSTM-LSTM models. Fig.~\ref{fig:modeltradeoff} (b) presents a 3D representation of the integrals of \(L_{\text{spectral}}\), \(D_{JS}\), and the average computation time per epoch across all architectures. In this multidimensional space, the LSTM-CNN model demonstrates a notable reduction in integral volume, indicative of its optimal balance between computational efficiency and loss minimization. These results suggest that the LSTM-CNN architecture optimally harmonizes the complexity of LSTM layers with the computational advantages afforded by CNN layers, establishing it as a robust candidate for high-fidelity synthetic data generation with minimized training overhead.
%%%%%%%%%%%%%%%%%%%%%%%%%%%%%%%%%%%%%%%%%%%%%%
%%%%%%%%%%%%%%%%%%%%%%%%%%%%%%%%%%%%%%%%%%%%%%

%%%%%%%%%%%%%%%%%%%%%%%%%%%%%%%%%%%%%%%%%%%%%%%%%%%%%%%%%%%%%%%%%%%%%%%%%%%%%%%%%%%%%%%%%%%%%%%%%%%%%%%%%%%%%%%%%%%%%%%%%%%%%%%%%%%%%%%%%%%%%%%%%%%%%%%%%%%%%%%%%%%%%%%%%%%%%%%%%%%%%%%%%%%%%%%%%%%%%%%%%%%%%%%%%%%%%%%%%%%%%%%%%%%%%%%%%%%%%%%%%%%%%%%%%%%%%%%%%%%%%%%%%%%%%%%%%%%%%%%%%%%%%%%%%%%%%%%%%%%%%%%%%%%%%%%%%%%%%%%%%%%%%%%%%%%%%%%%%%%%%%%%%%%%%%%%
\subsection{GAN versus Markov Model Experiment}
\begin{figure}[t]
    \centering
    \includegraphics[width=1\linewidth]{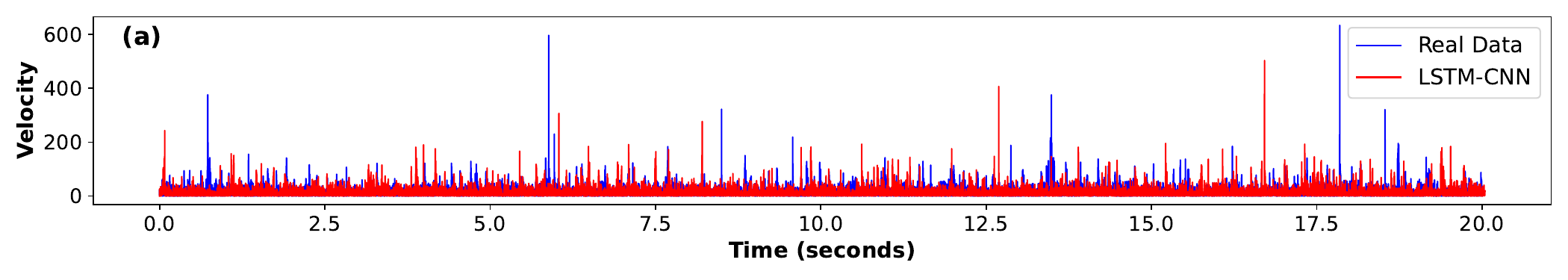}\\
    \includegraphics[width=0.495\linewidth]{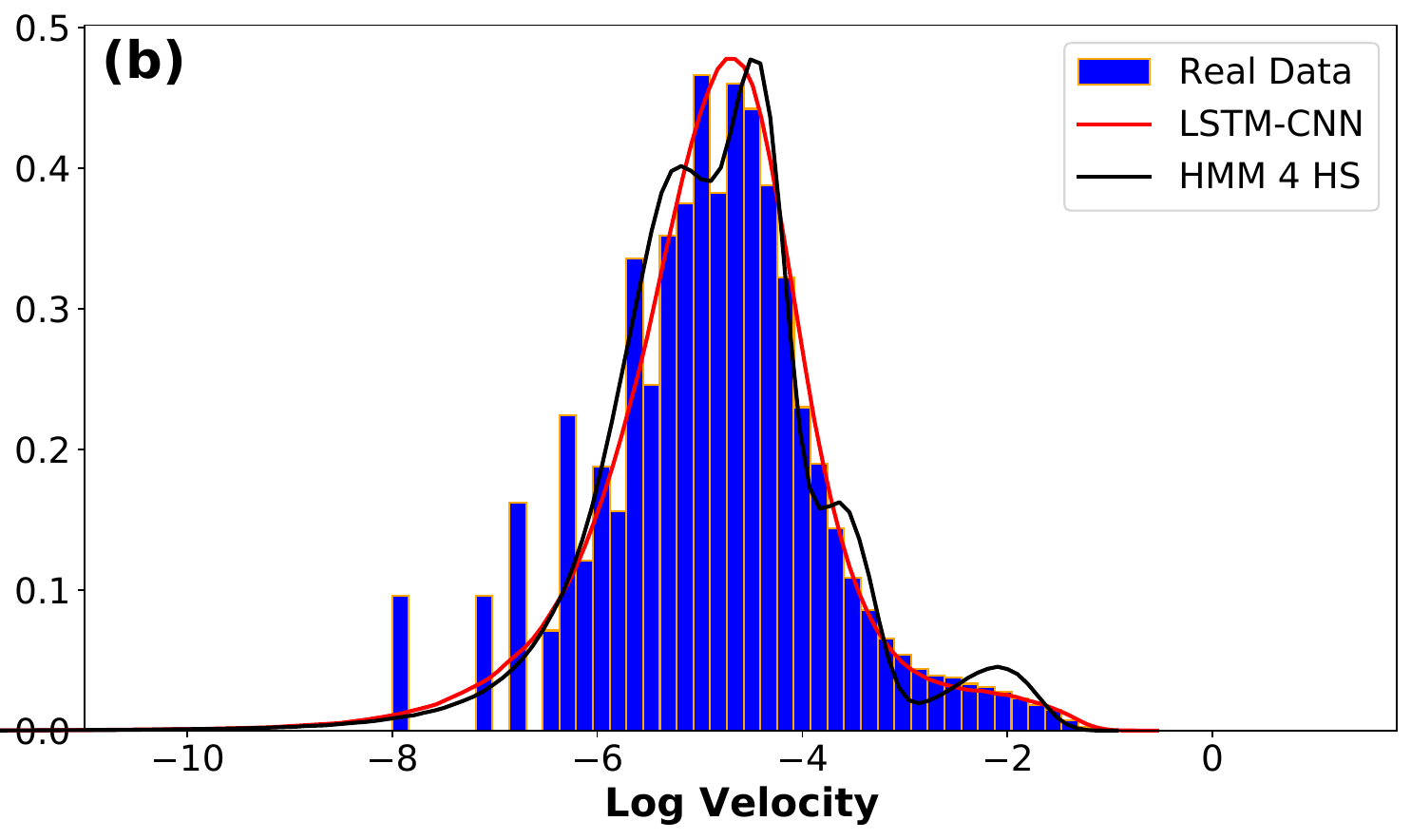}
    \includegraphics[width=0.495\linewidth]{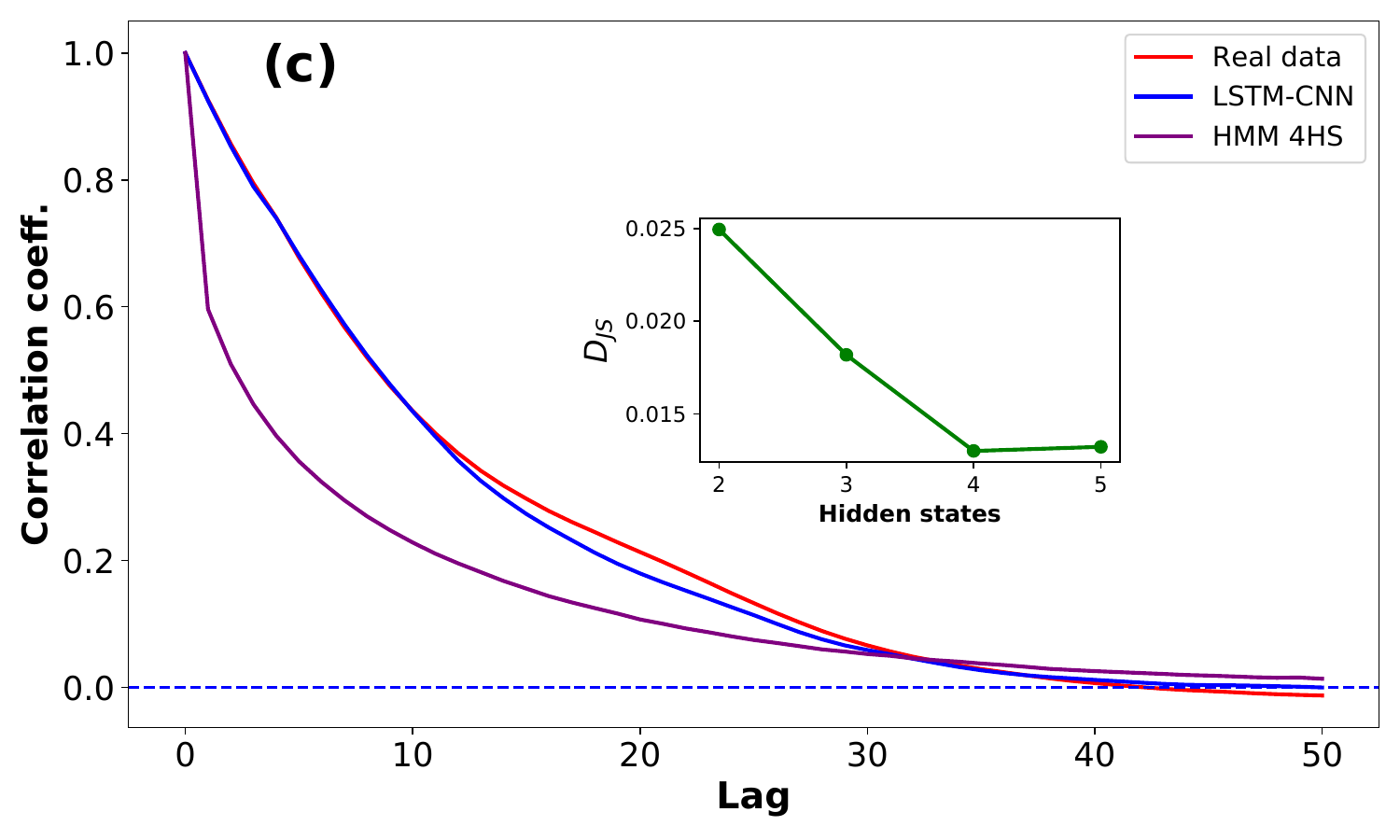}

    \caption{Comparison of real and generated data using GAN and HMM models. (a): The time series plot shows the differences between real and generated data over time for the LSTM-CNN model. (b): The histogram (blue) represents the real data distribution while the red and black curves represent the LSTM-CNN model with \( L_{\text{final}} \) and the HMM model respectively. The LSTM-CNN + \( L_{\text{final}} \) (red curve) shows closer alignment with the real data in terms of capturing the broader distribution, however, the HMM (black curve) is more concentrated, highlighting differences in model performance between the two. (c):  Autocorrelation plot for real data (red) and generated data from LSTM-CNN with $L_{\text{final}}$ (blue), and HMM with 4 hidden states (brown). The inset on the same plot shows JS divergence ($D_{JS}$) between real and HMM-generated data as a function of the number of hidden states (2 to 5).}
    \label{results}
    %\Description{Results}
\end{figure}

In this study, we compared the performance of the best GAN architecture (LSTM-CNN) with that of a HMM for modeling eye gaze velocity trajectories. The CNN-LSTM GAN utilized a random noise vector of 256 samples drawn from a uniform distribution as input. The models were trained on eye gaze velocity data consisting of 200-step time series sequences. The training was conducted over 500 epochs with a batch size of 128 and a learning rate of 0.0002. To enhance the GAN's capability to capture the spectral characteristics of eye gaze movements—particularly at fine temporal resolutions—and to stabilize the learning process, spectral regularization $L_{final}$ was implemented within the GAN framework. We applied an HMM to the same eye gaze velocity data for comparison. The HMM modeled sequences of stochastic eye gaze velocities with a transition matrix dictating state transitions, optimized via the Expectation-Maximization (EM) algorithm during model fitting. To determine the optimal number of hidden states, we evaluated models with two to five states by calculating $D_{JS}$ between the real data and the HMM-generated data, as shown in Fig.~\ref{results} (c). We observed that the $D_{JS}$ decreased from approximately 0.0249 with two states to 0.0131 with four states, but then increased slightly to 0.013245 with five states. Based on this analysis, we selected a model with four hidden states as a balance between model complexity and performance. The emission probability matrix related hidden states to observable eye gaze velocities, enabling the HMM to capture the dynamics of eye gaze trajectories effectively and allowing for a concise comparison with the GAN model.

%%%%%%%%%%%%%%%%%%%%%%%%%%%%%%%%%%%%%%%%%%%%%%%%%%%%%%%%%%%%%%%%%%%%%%%%%%%%%%%%%%%%%%%%%%%%%%%%%%%%%%%%%%%%%%%%%%%%%%%%%%%%%%%%%%%%%%%%%%%%%%%%%%%%%%%%%%%%%%%%%%%%%%%%%%%%%%%%

%%%%%%%%%%%%%%%%%%%%%%%%%%%%%%%%%%%%%%%%%%%%%%
%%%%%%%%%%%%%%%%%%%%%%%%%%%%%%%%%%%%%%%%%%%%%%
Table \ref{tab:stats_summary} compares the statistics for real and generated eye gaze velocity data LSTM-CNN and HMM. The LSTM-CNN model exhibits the closest match to the real data, with a mean of 0.00204 and a standard deviation of 0.0029, while also showing relatively minimal deviation in skewness (4.865) and kurtosis (34.204). This model achieves the lowest average loss (0.0471) and JS divergence (0.000322), indicating its ability to generate data that closely mimics the distribution of real eye gaze velocity patterns. On the other hand, the HMM-generated data demonstrates significant divergence from the real data, with a higher mean (0.0498) and a higher standard deviation (0.0730). The higher JS divergence (0.00714) further underscores the challenge the HMM faces in capturing the true complexity of the eye gaze velocity patterns. These results highlight the superior performance of the LSTM-CNN model in replicating real-world eye gaze dynamics.
%%%%%%%%%%%%%%%%%%%%%%%%%%%%%%%%%%%%%%%%%%%%%%
\begin{table}[t]
\caption{Comparison of real velocity data with GANs, and HMM statistics.}

\centering
\begin{tabular}{lccc}
\toprule
\textbf{Statistic} & \textbf{Real Data}  & \textbf{LSTM-CNN} & \textbf{HMM} \\
\midrule
Mean & 0.0021061  & 0.0020410 & 0.0498 \\
Standard Deviation & 0.0029530 &  0.0029640 & 0.0730 \\
Skewness & 4.7095 & 4.8659  & 4.39038\\
Kurtosis & 30.9307 &  34.2044 & 23.7580\\
\midrule
\textbf{Average $L_{final}$ score} & -- &  0.0471601 & -- \\
\textbf{Average $D_{JS}$ score} & -- &  \textbf{0.0003222} & 0.01244\\
\bottomrule
\end{tabular}
\label{tab:stats_summary}
\end{table}
%%%%%%%%%%%%%%%%%%%%%%%%%%%%%%%%%%%%%%%%%%%%%%

Fig.~\ref{results} presents a comparative analysis of real and generated data distributions using GAN (LSTM-CNN) and HMM approaches. The figure showcases how each model captures the characteristics of the real data and provides insight into their strengths and limitations across multiple performance metrics. In Fig.~\ref{results} (a), we observe the log-scaled velocity distributions for real data, the LSTM-CNN model, and the HMM model. The LSTM-CNN model shows a closer alignment with the real data distribution, indicating that incorporating spectral loss helps capture more variability and nuances in the data. The HMM model with four hidden states shows a relatively good fit to the real data, although some deviations are visible. These deviations indicate that while the HMM can model certain aspects of the real data, it may still miss some of the finer details, such as the distribution's tails, which the LSTM-CNN better captures with a spectral loss plot. Fig.~\ref{results} (b) shows autocorrelation coefficients for the real data and the generated data from two different models: LSTM-CNN, and HMM with four hidden states. LSTM-CNN models capture the overall trend in the autocorrelation function. The HMM-generated data shows more pronounced deviations, particularly in the early lags, suggesting that the HMM model struggles to replicate the temporal dependencies present in the real data. The inset in the same plot displays the \(D_{JS}\) between real and HMM-generated data as a function of the number of hidden states (from 2 to 5). %As the number of hidden states increases from 2 to 4, the \(D_{JS}\) decreases, indicating that the model's performance improves with more hidden states up to four. However, when increasing to five hidden states, the \(D_{JS}\) slightly decreases, suggesting that adding more states beyond four does not further enhance the model's performance. 

%%%%%%%%%%%%%%%%%%%%%%%%%%%%%%%%%%%%%%%%%%%%%%%%%%%%%%%%%%%%%%%%%%%%%%%%%%%%%%%%%%%%%%%%%%%%%%%%%%%%%%%%%%%%%%%%%%%%%%%%DISCUSSION%%%%%%%%%%%%%%%%%%%%%%%%%%%%%%%%%%%%%%%%%%%%%%%%%%%%%%%%

\section{Discussion}

The accurate modeling of eye gaze velocity trajectories is crucial for advancements in fields such as human-computer interaction, neuropsychology, and cognitive science. These trajectories exhibit complex temporal dynamics and spectral properties due to rapid and subtle eye moments, making them challenging to model accurately. In this study, we conducted a comprehensive evaluation of various GAN architectures, specifically CNN-CNN, LSTM-CNN, CNN-LSTM, and LSTM-LSTM, augmented with spectral loss (\(L_{\text{final}}\)), and compared their performance with Markov Model. Our findings indicate that the LSTM-CNN GAN architecture, when trained with spectral loss, significantly outperforms the Markov model in capturing the complex temporal and distributional characteristics of eye gaze velocity data. The LSTM-CNN model trained achieved the lowest Jensen-Shannon divergence (\(D_{JS} = 0.00032\)), indicating an exceptional alignment with the real data distribution. This superior performance can be attributed to the architectural synergy where LSTM layers in the generator effectively model long-term temporal dependencies, and CNN layers in the discriminator capture local spatial patterns. The incorporation of spectral loss emphasizes the frequency components of the data, ensuring that both high-frequency and low-frequency elements are accurately modeled. The spectral loss not only improved the fidelity of the generated data but also contributed to a more stable and efficient training process. This is evidenced by the reduced integral volume in the three-dimensional evaluation space (Fig.~\ref{fig:modeltradeoff}), indicating consistent performance across multiple evaluation metrics. 
%HMM model, optimized with four hidden states to balance complexity and performance, demonstrated a higher \(D_{JS}\) of 0.004071.
While HMMs are adept at modeling sequences with clear state transitions, they are limited in capturing the intricate, non-linear temporal dynamics inherent in eye gaze trajectories. The statistical discrepancies observed—such as the significant deviation in mean and standard deviation (Table~\ref{tab:stats_summary})—underscore the HMM's limitations in replicating the nuanced patterns of eye movement data. Our results contrast with the previous research highlighting the challenges faced by traditional GANs and HMMs in modeling complex time-series data. For instance, Lencastre et al. (2023) \cite{LENCASTRE2023133831} reported that conventional GANs often struggle with capturing rare events and maintaining time continuity, leading to inadequate modeling of the distribution tails and temporal dependencies. Their study found Markov models to outperform GANs in replicating statistical moments and cross-feature relationships. However, our findings demonstrate that when enhanced with spectral regularization and an appropriate architecture design, GANs—specifically the LSTM-CNN model—can surpass the performance of the Markov model, effectively capturing both the global distribution and the temporal autocorrelation structures of eye gaze data.

Furthermore, the study by Bhandari et al. (2024) \cite{10.1145/3638530.3664134} explored the use of Quantum GANs (QGANs) for synthetic data generation. Despite the theoretical advantages of quantum computing in handling complex probability distributions, their results indicated a higher \(D_{JS}\) compared to classical GANs, suggesting that QGANs are currently less effective in accurately replicating real-world data distributions. This disparity could be due to the nascent stage of quantum computing technologies and the challenges associated with implementing quantum algorithms for practical data modeling tasks.

The superior performance of the LSTM-CNN GAN with spectral loss in our study highlights several key insights: 1) architectural synergy: The combination of LSTM layers in the generator with CNN layers in the discriminator leverages the strengths of both architectures, enabling the model to capture long-term dependencies and local features effectively. 2) Spectral loss: Incorporating spectral loss helps the GAN focus on the frequency domain characteristics of the data, ensuring that both high-frequency and low-frequency components are accurately modeled. This is particularly important for eye gaze data, which exhibits complex spectral properties due to rapid and subtle movements. 3) Computational efficiency: Despite the increased complexity, the LSTM-CNN GAN maintains computational efficiency, balancing training overhead with performance gains. This makes it a practical choice for applications requiring real-time or near-real-time data generation.  
%%%%%%%%%%%%%%%%%%%%%%%%%%%%%%%%%%%%%%%%%%%%%%%%%%%%%%%%%%%%%%%%%%%%%%%%%%%%%%%%%%%%%%%%%%%%%%%%%%%%%%%%%%%%%%%%%CONCLUSION%%%%%%%%%%%%%%%%%%%%%%%%%%%%%%%%%%%%%%%%%%%%%%%%%%%%%%%%%%%%%%%%%%%%%%%
\section{Conclusion}

Our study demonstrates the potential of spectrally regularized LSTM-CNN GANs in generating high-fidelity synthetic eye gaze velocity trajectories. By effectively capturing the intricate temporal and spectral characteristics of real eye gaze data, the LSTM-CNN GAN outperforms the HMMs. It addresses some of the limitations identified in previous studies involving GANs and QGANs. These findings have significant implications for the development of simulation environments, training systems, and eye-tracking technologies that rely on realistic synthetic data. The results contribute to the broader goal of enhancing human-computer interaction by providing models that simulate naturalistic eye movement behaviors. As GAN architectures and training methodologies continue to evolve, their application in modeling complex biological signals holds promise for research and practical implementations across various domains.

%%%%%%%%%%%%%%%%%%%%%%%%%%%%%%%%%%%%%%%%%%%%%%%%%%%%%%%%%%%%%%%%%%%%%%%%%%%%%%%%%%%%%%%%%%%%%%%%%%%%%%%%%%%%%%%%%%%%%%%%%%%%%%%%%%%%%%%%%%%%%%%%%%%%%%%%%%%%%%%%%%%%%%%%%%%%%%%%%%%%%%%%
\section*{Author Contributions Statement}

S.B. conducted the primary research, performed the experiments, carried out simulations, and led the analysis and the main writing of the manuscript. 
P.L. supervised the data collection project and developed and implementated the Markov model.
R.M. collected and curated the data.  
A.S. and A.Y. contributed to the theoretical analysis and background of the results from GANs and the Markovian approaches. 
P.G.L. coordinated the overall project. All authors contributed to the analysis and discussion of the results as well as the writing and revision of the manuscript. 
%Additionally, all authors revised the text.
%%%%%%%%%%%%%%%%%%%%%%%%%%%%%%%%%%%%%%%%%%%%%%%%%%%%%%%%%%%%%%%%%%%%%%%%%%%%%%%%%%%%%%%%
\section*{Data Availability}
The datasets analyzed during the current study are available from the corresponding author upon reasonable request.
%%%%%%%%%%%%%%%%%%%%%%%%%%%%%%%%%%%%%%%%%%%%%%%%%%%%%%%%%%%%%%%%%%%%%%%%%%%%%%%%%%%%%%%%
\section*{Code Availability}
 The code and the generated data used during and/or analyzed during the current study are available in the GANsForVirtualEye repository (GitHub: \url{https://github.com/shailendrabhandari/GANsForVirtualEye.git}) or as a Python package (PyPI: \url{https://pypi.org/project/GANsforVirtualEye/}).
%%%%%%%%%%%%%%%%%%%%%%%%%%%%%%%%%%%%%%%%%%%%%%%%%%%%%%%%%%%%%%%%%%%%%%%%%%%%%%%%%%%%%%%%
\section*{Acknowledgments}
This work is funded by the Research Council of Norway under grant number 335940 for the project `Virtual-Eye'.

%%%%%%%%%%%%%%%%%%%%%%%%%%%%%%%%%%%%%%%%%%%%%%%%%%%%%%%%%%%%%%%%%%%%%%%%%%%%%%%%%%%%%%%%%%%%%%%%%%%%%%%%%%%%%%%%%%%%%%%%%%%%%%%%%%%%%%%%%%%%%%%%%%%%%%%%%%%%%%%%%%%%%%%%%%%%%%%%%%%%%%%%%%%%%%%%%%%%%%%%%%%%%%%%%%%%%%%%%%%%%%%%%%%%%%%%%%%%%%%%%%%%%%%%%%%%%%%%%%%%%%%%

\bibliography{main}

%%%%%%%%%%%%%%%%%%%%%%%%%%%%%%%%%%%%%%%%%%%%%%%%%%%%%%%%%%%%%%%%%%%%%%%%%%%%%%%%%%%%%%%%%%%%%%%%%%%%%%%%%%%%%%%%%%%%%%%%%APPENDIX%%%%%%%%%%%%%%%%%%%%%%%%%%%%%%%%%%%%%%%%%%%%%%%%%%%%%%%%%%%%%%%%%%%%%%%%%%%%%%%%%%%%%%%%%%%%%%%%%%%%%%%%%%%%%%%%%%%%%%%%%%%%%%%%%%%%%%%%%%%%%%%%%%%%%%%%%%%%%%%%%%%%%%%%%%%%%%%%%%%%%%%%%%%%%%%%%%%%%%%%%%%%%%%%%%%%%%%%%%%%%%%%%%%%%%%
\appendix
\section{Mathematical Foundation for Markov and Hidden Markov Models}\label{appexA}
A sequence of random variables \( \{X_n\}_{n \geq 0} \) with values in a set \( E \) is known as a discrete-time stochastic process with state space \( E \). In this context, the state space is assumed to be countable, and its elements are denoted by \( i, j, k, \ldots \). If \( X_n = i \), we say that the process is in state \( i \) at time \( n \), or that it visits state \( i \) at time \( n \).

\textbf{Definition: the Markov property}: Let \( \{X_n\}_{n \geq 0} \) represent a discrete-time stochastic process with a countable state space \( E \). If for any integer \( n \geq 0 \) and for all states \( i_0, i_1, \ldots, i_{n-1}, i, j \),
\begin{equation}
    P(X_{n+1} = j \mid X_n = i, X_{n-1} = i_{n-1}, \ldots, X_0 = i_0) = P(X_{n+1} = j \mid X_n = i),
\end{equation}
whenever both sides are well-defined, this stochastic process is referred to as a Markov chain \cite{Bremaud2020}. If, in addition, the right-hand side of this expression is independent of \( n \), the chain is called a \textit{homogeneous Markov chain (HMC)}.

The matrix \( P = \{p_{ij}\}_{i,j \in E} \), where
\begin{equation}
    p_{ij} = P(X_{n+1} = j \mid X_n = i),
\end{equation}
is known as the \textit{transition matrix} of the HMC. As its entries are probabilities, and since a transition from any state \( i \) must lead to some state, it follows that
\begin{equation}
    p_{ij} \geq 0, \quad \sum_{k \in E} p_{ik} = 1
\end{equation}
for all states \( i, j \). A matrix \( P \) indexed by \( E \) and satisfying these properties is called a \textit{stochastic matrix}.
For a \( k \)-th order Markov process, the conditional probability can be estimated via Kernel Density Estimation (KDE) as:
\begin{equation}
    P(X_{n} \mid X_{n-1}^{n-k}) = \frac{f(X_{n}, X_{n-1}^{n-k})}{f(X_{n-1}^{n-k})},
\end{equation}
where \( X_{n-1}^{n-k} = \{X_{n-1}, X_{n-2}, \ldots, X_{n-k}\} \) and the joint probability density function \( f(\cdot) \) is estimated by:
\begin{equation}
    f(X_{n}, X_{n-1}^{n-k}) = \frac{1}{(N - k) h^{k+1}} \sum_{i=k+1}^{N} \prod_{j=0}^{k} K\left( \frac{X_{n-j} - X_{i-j}}{h} \right),
\end{equation}
where \( K(u) \) is the Gaussian kernel function:
\begin{equation}
    K(u) = \frac{1}{\sqrt{2\pi}} \exp\left( -\frac{u^2}{2} \right),
\end{equation}
and the bandwidth \( h \) is calculated using Silverman's rule \cite{silverman1998density}:
\begin{equation}
    h = 1.06 \hat{\sigma} N^{-1/(k+4)},
\end{equation}
where \( \hat{\sigma} \) represents the standard deviation of the data.

\begin{figure}[t]
    \includegraphics[width=0.49\linewidth]{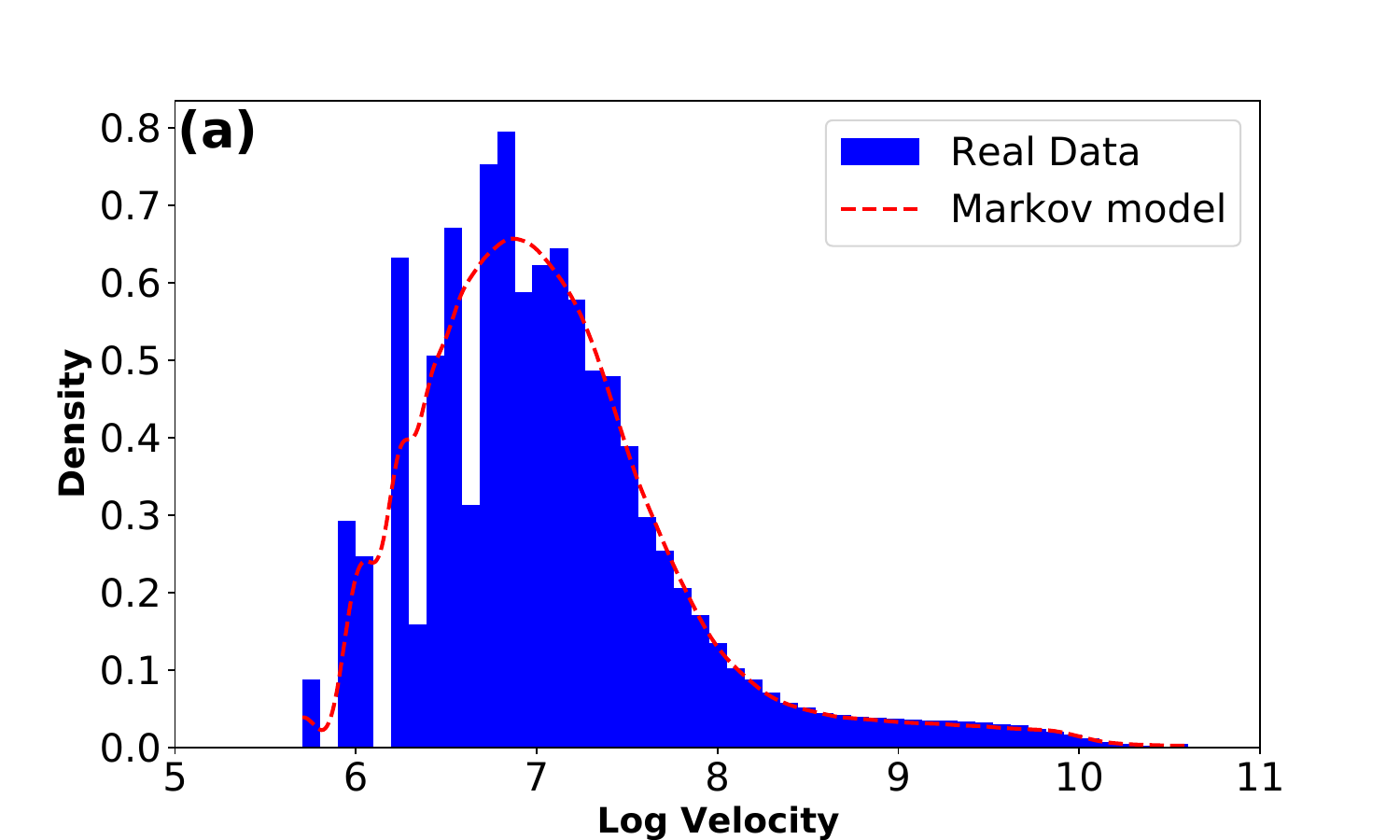}
    \includegraphics[width=0.49\linewidth]{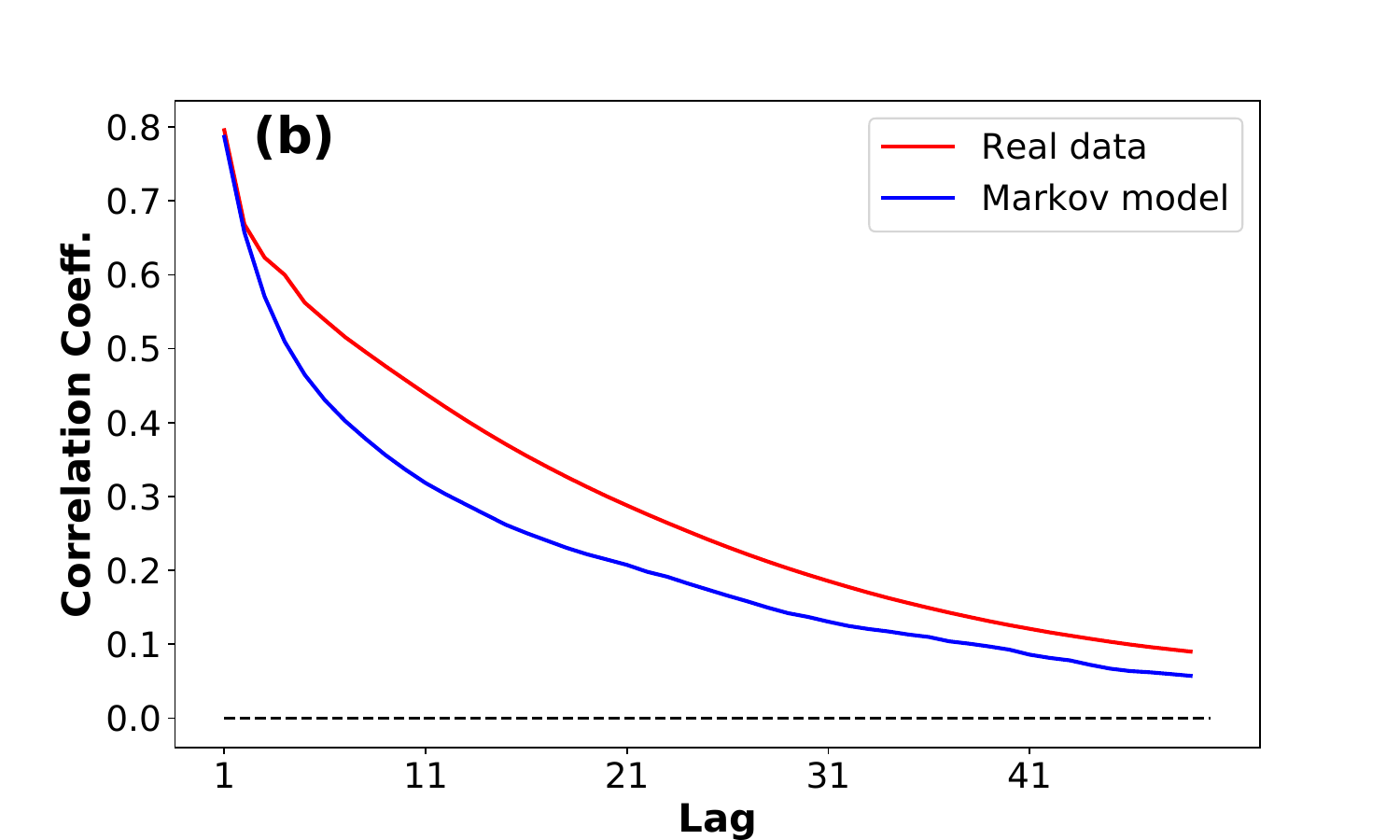}
    \caption{(a) Histogram of log-transformed velocities in real data with the density estimation from a Markov model, illustrating the similarity in velocity distributions. The real data (blue bars) aligns closely with the Markov model’s Kernel Density Estimation (KDE) curve (dashed red line), suggesting that the synthetic data captures key distributional features. (b) Autocorrelation plots of real (red line) and synthetic (blue line) data over varying lag intervals, show a faster decrease in correlation for the synthetic data compared to real data. This indicates that while the synthetic data replicates short-term dependencies, it may diverge in capturing long-term dependencies observed in the real data.}
    \label{results_markov}
    %\Description{App Results.}
\end{figure}

To evaluate the effectiveness of the Markov model in capturing the statistical properties of eye-gaze velocities, we analyze the real and the generated data distributions and their autocorrelation. Figure \ref{results_markov} presents these comparisons in detail. Fig. \ref{results_markov} (a), displays the histogram of log-transformed velocities from the real data alongside the velocity eye gaze trajectories in the log scale derived from the Markov model. The close alignment between the blue bars (real data) and the dashed red line (Markov model) indicates that the synthetic data generated by the Markov model successfully captures the key distributional features of the real data. Fig. \ref{results_markov} (b) shows the autocorrelation plots of both real (red line) and synthetic (blue line) data over varying lag intervals. The generated data exhibits a faster decrease in autocorrelation compared to the real data, suggesting that while the Markov model replicates the short-term dependencies effectively, it may not fully capture the long-term dependencies present in the real data. This observation highlights a limitation of the Markov model in modeling temporal dependencies over longer time scales in eye-gaze velocity trajectories.

Hidden Markov Models (HMMs) are powerful statistical tools for modeling sequential data where the system being modeled is assumed to be a Markov process with unobserved (hidden) states \cite{rabiner1989tutorial}. They are beneficial in scenarios where we can observe a sequence of emissions (observable events) probabilistically dependent on a sequence of hidden states that are not directly observable. Several components formally define an HMM. First, the hidden states \( S = \{s_1, s_2, \dots, s_N\} \), where \( N \) is the number of hidden states, each representing a distinct mode of the system, such as different types of eye movements. Next, the observations \( O = \{o_1, o_2, \dots, o_T\} \), where \( T \) is the length of the observation sequence, with each \( o_t \) representing the observed eye-gaze velocity at time \( t \). The initial state distribution is denoted as \( \boldsymbol{\pi} = \{\pi_i\} \), where \( \pi_i = P(q_1 = s_i) \) is the probability that the Markov chain starts in the state \( s_i \). The state transition probability matrix is given by \( A = [a_{ij}] \), where \( a_{ij} = P(q_{t+1} = s_j \mid q_t = s_i) \) represents the probability of transitioning from state \( s_i \) to state \( s_j \). Finally, the emission probability distribution is represented as \( B = \{b_j(o_t)\} \), where \( b_j(o_t) = P(o_t \mid q_t = s_j) \) is the probability of observing \( o_t \) given that the system is in state \( s_j \) at time \( t \). The complete parameter set of the HMM is denoted as \( \lambda = (A, B, \boldsymbol{\pi}) \).
The joint probability of observing a sequence \( O \) and a particular state sequence \( Q = \{q_1, q_2, \dots, q_T\} \) is expressed as:
\begin{equation}
    P(O, Q \mid \lambda) = \pi_{q_1} b_{q_1}(o_1) \prod_{t=2}^{T} a_{q_{t-1} q_t} b_{q_t}(o_t),
\end{equation}
where ($\lambda=\pi,A,B$ ) represents the parameters of the HMM.
Since the states \( Q \) are hidden, the focus shifts to computing the likelihood of the observations \( O \), which is given by:
\begin{equation}
    P(O \mid \lambda) = \sum_{Q} P(O, Q \mid \lambda).
\end{equation}
However, directly computing this sum is computationally infeasible for large \( T \), as it involves summing over \( N^T \) possible state sequences. To overcome this challenge, efficient algorithms like the Forward-Backward algorithm \cite{rabiner1989tutorial} are employed. We employed the Baum-Welch algorithm, a specialized instance of the Expectation-Maximization (EM) algorithm tailored for HMMs, to estimate the model parameters \(\lambda = (\pi, A, B)\). This algorithm iteratively refines the estimates of the initial state probabilities \(\pi\), the state transition probabilities \(A\), and the emission probabilities \(B\) to maximize the likelihood of the observed data \cite{baum1970maximization}. In the initialization phase, we start with initial guesses for \(\pi\), \(A\), and \(B\). The algorithm then proceeds through iterative Expectation and Maximization steps until convergence. The expectation step occupancies and expected state transition counts are computed using the forward-backward procedure. The forward probabilities \(\alpha_t(i)\) and backward probabilities \(\beta_t(i)\) are calculated recursively to evaluate the likelihood of partial observation sequences \cite{YU2016143}.
\begin{equation}
    \alpha_1(i) = \pi_i \, b_i(o_1), \quad \alpha_{t+1}(j) = \left( \sum_{i=1}^N \alpha_t(i) \, a_{ij} \right) b_j(o_{t+1}),
\end{equation}
where \( \alpha_t(i) \) is the probability of observing the partial sequence \( o_1, o_2, \dots, o_t \) and being in state \( s_i \) at time \( t \).
\begin{equation}
    \beta_T(i) = 1, \quad \beta_t(i) = \sum_{j=1}^N a_{ij} \, b_j(o_{t+1}) \, \beta_{t+1}(j), b_j(o_{t+1}),
\end{equation}
where \( \beta_t(i) \) is the probability of observing the partial sequence \( o_{t+1}, o_{t+2}, \dots, o_T \) given that the state at time \( t \) is \( s_i \).
Using these probabilities, we calculate the expected state occupancy \(\gamma_t(i)\) and the expected state transitions \(\xi_t(i, j)\):
\begin{equation}
    \gamma_t(i) = \frac{\alpha_t(i) \beta_t(i)}{\sum_{k=1}^N \alpha_t(k) \beta_t(k)},
\end{equation}
which represents the probability of being in state \( s_i \) at time \( t \) given the observed sequence.
 \begin{equation}
     \xi_t(i, j) = \frac{\alpha_t(i) \, a_{ij} \, b_j(o_{t+1}) \, \beta_{t+1}(j)}{\sum_{i=1}^N \sum_{j=1}^N \alpha_t(i) \, a_{ij} \, b_j(o_{t+1}) \, \beta_{t+1}(j)},
 \end{equation}
which represents the probability of transitioning from state \( s_i \) at time \( t \) to state \( s_j \) at time \( t+1 \) given the observed sequence. We update the parameters to maximize the expected log-likelihood calculated in the E-step: $\pi_i = \gamma_1(i)$, ensuring that the estimated initial state probabilities reflect the expected occupancy at time \( t = 1 \).
\begin{equation}
     a_{ij} = \frac{\sum_{t=1}^{T-1} \xi_t(i, j)}{\sum_{t=1}^{T-1} \gamma_t(i)},
\end{equation}
where the numerator is the expected number of transitions from state \( s_i \) to \( s_j \), and the denominator is the expected number of transitions from state \( s_i \). For a Gaussian emission model with mean \( \mu_i \) and variance \( \sigma_i^2 \), the parameters are updated as:

\begin{equation}
    \mu_i = \frac{\sum_{t=1}^T \gamma_t(i) \, o_t}{\sum_{t=1}^T \gamma_t(i)}, \quad \text{and}\quad \sigma_i^2 = \frac{\sum_{t=1}^T \gamma_t(i) \left( o_t - \mu_i \right)^2}{\sum_{t=1}^T \gamma_t(i)},
\end{equation}
where \( o_t \) is the observed value at time \( t \).
These expectation and maximization steps are iterated until convergence criteria are met, typically when the increase in log-likelihood between iterations falls below a predefined threshold or after reaching a maximum number of iterations. This iterative process ensures that the parameter estimates progressively improve, leading to a model that effectively captures the underlying dynamics of the eye gaze velocity trajectories. By employing the Baum-Welch algorithm, we leveraged a robust statistical framework for parameter estimation in HMMs, which is well-suited for modeling time series data with hidden structures \cite{rabiner1989tutorial}. This approach allowed us to systematically infer the hidden states and optimize the model parameters based on the observed data, facilitating a comprehensive analysis of eye movement behaviors.

\end{document}